\documentclass[letterpaper, 10 pt, conference]{ieeeconf}  

\IEEEoverridecommandlockouts                              
\usepackage{amsmath,amssymb,stmaryrd,mathtools}
\usepackage{amsfonts}

\usepackage[noend]{algpseudocode}
\usepackage{acronym}
\usepackage{verbatim}
\usepackage{booktabs}
\usepackage{siunitx}
\usepackage{graphics}
\usepackage{graphicx,caption}
\usepackage{flushend}
\usepackage{suffix}
\usepackage{xstring}
\usepackage{xparse}
\usepackage{expl3}
\usepackage{mathrsfs}
\usepackage{tabularx}
\usepackage{makecell}
\usepackage{array}
\usepackage[hidelinks]{hyperref}

\usepackage{cleveref}
\usepackage[ruled,linesnumbered]{algorithm2e}
\usepackage{multirow}
\usepackage{diagbox}
\usepackage{rotating}

\usepackage{subcaption}
\usepackage{wrapfig}

\usepackage{tikz}
\usetikzlibrary{arrows,backgrounds,calc}
\usepackage{relsize}
\usepackage{float}
\usepackage{kantlipsum} 
\usepackage{lipsum}
\usepackage{stfloats}
\usepackage{siunitx}

\usepackage[noadjust]{cite}
\usepackage{todonotes}
\usepackage{soul}
\definecolor{smoothgreen}{rgb}{0.7,1,0.7}
\sethlcolor{smoothgreen}

\RequirePackage{luatex85}
\usepackage{pgfplots}
\pgfplotsset{compat=newest}
\pgfplotsset{every axis legend/.append style={%
		cells={anchor=west}}
}
\usepgfplotslibrary{polar}
\usetikzlibrary{arrows}
\tikzset{>=stealth'}

\definecolor{C1}{rgb}{0.0, 0.447, 0.741}
\definecolor{C1_light}{rgb}{0.0, 0.6032388663967612, 1.0}
\definecolor{C2}{rgb}{0.85, 0.325, 0.098}
\definecolor{C3}{rgb}{0.929, 0.694, 0.125}
\definecolor{C4}{rgb}{0.494, 0.184, 0.556}
\definecolor{C5}{rgb}{0.466, 0.674, 0.188}
\definecolor{C6}{rgb}{0.301, 0.745, 0.933}
\definecolor{C7}{rgb}{0.635, 0.078, 0.184}

\usepgfplotslibrary{groupplots}

\usetikzlibrary{shapes.geometric, arrows}

\tikzstyle{startstop} = [rectangle, rounded corners, minimum width=2cm, minimum height=1cm,text centered, draw=black, fill=none]
\tikzstyle{arrow} = [thick,->,>=stealth]

\title{
Decentralized Structural-RNN for Robot Crowd Navigation\\with Deep Reinforcement Learning
}

\author{%
Shuijing Liu*, Peixin Chang*, Weihang Liang$\dagger$, \\ Neeloy Chakraborty$\dagger$, and Katherine Driggs-Campbell%
\thanks{*,$\dagger$ denote equal contribution.}%
\thanks{S. Liu, P. Chang, W. Liang, N. Chakraborty and K. Driggs-Campbell are with the Department of  Electrical and Computer Engineering at the University of Illinois at Urbana-Champaign. emails: \{sliu105,pchang17,weihang2,neeloyc2,krdc\}@illinois.edu}%
}

\begin{document}
\maketitle
\thispagestyle{empty}
\pagestyle{empty}

\begin{abstract}
Safe and efficient navigation through human crowds is an essential capability for mobile robots. 
Previous work on robot crowd navigation assumes that the dynamics of all agents are known and well-defined. In addition, the performance of previous methods deteriorates in partially observable environments and environments with dense crowds. 
To tackle these problems, we propose decentralized structural-Recurrent Neural Network (DS-RNN), a novel network that reasons about spatial and temporal relationships for robot decision making in crowd navigation. 
We train our network with model-free deep reinforcement learning without any expert supervision. 
We demonstrate that our model outperforms previous methods in challenging crowd navigation scenarios. We successfully transfer the policy learned in the simulator to a real-world TurtleBot 2i.
For more information, please visit the project website at {\color{cyan}\url{https://sites.google.com/view/crowdnav-ds-rnn/home}}.
\end{abstract}

\section{Introduction}
\label{sec:intro}
As mobile robots are becoming prevalent in people's daily lives, autonomous navigation in crowded places with other dynamic agents is an important yet challenging problem~\cite{kruse2013human,du2019online}. 
Inspired by the recent applications of deep learning in robot control~\cite{mnih2015human,levine2016end,gupta2017cognitive,chang2020robot} and in graph modeling~\cite{jain2016structural}, we seek to build a learning-based graphical model for mobile robot navigation in pedestrian-rich environments.

Robot crowd navigation is a challenging task for two key reasons. First, the problem is decentralized, meaning that each agent runs its own policy individually, which makes the environment not fully observable to the robot. For example, other agents' preferred walking style and intended goals are not known in advance and are difficult to infer online~\cite{huang2020intention}. Second, the crowded environment contains both dynamic and static agents, who implicitly interact with each other during navigation.
The ways agents influence each other are often difficult to model~\cite{gupta2018social}, making the dynamic environment harder to navigate and likely to produce emergent phenomenon~\cite{bhattacharyya2019simulating}.

Despite these challenges, robot crowd navigation is well-studied and has had many successful demonstrations~\cite{fox1997dynamic,van2008reciprocal,chen2019crowd}. Reaction-based methods such as Optimal Reciprocal Collision Avoidance (ORCA) and Social Force (SF) use one-step interaction rules to determine the robot's optimal action~\cite{van2011reciprocal, van2008reciprocal,helbing1995social}. Another line of works first predict other agents' future trajectories and then plan a path for the robot~\cite{aoude2013probabilistically,kretzschmar2016socially,trautman2013robot,kuderer2012feature}. However, these two methods suffer from the \emph{freezing robot problem}: in dense crowds, the planner decides that all paths are unsafe and the robot freezes, which is suboptimal as a feasible path usually exists~\cite{trautman2010unfreezing}.

More recently, learning-based methods model the robot crowd navigation as a Markov Decision Process (MDP) and use Deep V-Learning to solve the MDP~\cite{chen2017decentralized,chen2017socially,everett2018motion,chen2019crowd,chen2020robot_gaze}. In Deep V-Learning, the agent chooses an action based on the state value approximated by neural networks. 
However, Deep V-Learning is typically initialized by ORCA using supervised learning and, as a result, the final policy inherits ORCA's aforementioned problems. Moreover, to choose actions from the value network, the dynamics of the humans are assumed to be known to the robot and are deterministic, which can be unrealistic in real applications.

\begin{figure}
    \centering
    \includegraphics[scale=0.2]{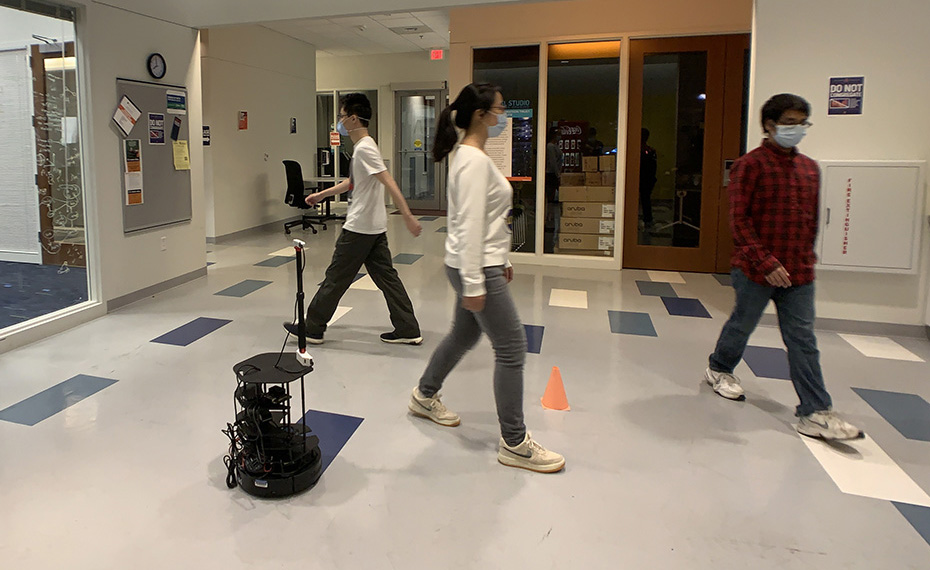}
    \caption{\textbf{Real-world crowd navigation with a TurtleBot 2i.} The orange cone on the floor denotes the robot goal. The TurtleBot is equipped with cameras for localization and human tracking.}
    \label{fig:opening}
    \vspace{-20pt}
\end{figure}

In this paper, we seek to create a learning framework for robot crowd navigation using spatio-temporal reasoning trained with model-free deep reinforcement learning (RL). 
We model the crowd navigation scenario as a decentralized spatio-temporal graph (st-graph) to capture the interactions between the robot and multiple humans through both space and time.
Then, we convert the decentralized st-graph to a novel end-to-end decentralized structural-RNN (DS-RNN) network. 
Using model-free RL, our method directly learns a navigation policy without prior knowledge of any agent's dynamics or expert policies. Since the robot learns entirely from its own experience, the resulting navigation policy easily adapts to dense human crowds and partial observability and outperforms previous methods in these scenarios. 

We present the following contributions: (1) We propose a novel deep neural network architecture called DS-RNN, which enables the robot to perform efficient spatio-temporal reasoning in crowd navigation; (2) We train the network using model-free RL without any supervision, which both simplifies the learning pipeline and avoids the network from converging to a suboptimal policy too early; (3) Our method demonstrates better performance in challenging navigation settings compared with previous methods. For more details, our code is available at \url{https://github.com/Shuijing725/CrowdNav_DSRNN}.

This paper is organized as follows: We review previous related works in Section~\ref{sec:related}. We formalize the problem and propose our network architecture in Section~\ref{sec:methods}. Experiments and results in simulation and in the real world are discussed in Section~\ref{sec:sim_exp} and Section~\ref{sec:real_exp}, respectively. Finally, we conclude the paper in Section~\ref{sec:conclusion}.

\section{Related Works}
\label{sec:related}


\subsection{Reaction-based methods}
\label{sec:reaction}

Robot navigation in dynamic environments has been studied for over two decades~\cite{fox1997dynamic,roy1999coastal, hoy2015algorithms, savkin2014seeking}. A subset of these works specifically focuses on robot navigation in pedestrian-rich environments or crowd navigation~\cite{kruse2013human,trautman2015robot}. 

Reaction-based methods such as Reciprocal Velocity Obstacle (RVO) and ORCA model other agents as velocity obstacles to find optimal collision-free velocities under reciprocal assumption~\cite{snape2011hybrid, van2011reciprocal, van2008reciprocal}. 
Another method named Social Force models the interactions in crowds using attractive and repulsive forces~\cite{helbing1995social}. However, these algorithms suffer from the \emph{freezing robot problem}~\cite{trautman2010unfreezing}. In addition, since the robot only uses the current states as input, the generated paths are often shortsighted and unnatural. 

In contrast, we train our network with model-free RL to mitigate the freezing problem. Also, our network contains RNNs that take a sequence of trajectories as input to encourage longsighted behaviors. 

\subsection{Trajectory-based methods}
\label{sec:trajectory}
Trajectory-based methods predict other agents' intended trajectories to plan a feasible path for the robot~\cite{aoude2013probabilistically,kretzschmar2016socially,trautman2013robot,kuderer2012feature,chen2019relational,eiffert2020path,cao2019dynamic}. Trajectory predictions allow the robot planner to look into the future and make long-sighted decisions. However, these methods have the following disadvantages. First, predicting trajectory sequences and searching a path from a large state space online are computationally expensive and can be slow in real time~\cite{driggs2017integrating}. 
Second, the predicted trajectories can make a large portion of the space untraversable, which might make the robot overly conservative~\cite{driggs2018robust}.

\subsection{Learning-based methods}
\label{sec:learning}
With the recent advancement of deep learning, imitation learning has been used to uncover policies from demonstrations of desired behaviors~\cite{tai2018socially,long2017deep}.
Another line of works use Deep V-Learning, which combines supervised learning and RL~\cite{chen2017decentralized,chen2017socially,everett2018motion,chen2019crowd,chen2019relational,chen2020robot_gaze}. 
Given the state transitions of all agents, the planner first calculates the values of all possible next states from a value network. Then, the planner chooses an action that leads to the state with the highest value.
To train the value network, Deep V-Learning first initializes the network by supervised learning using trajectories generated by ORCA, and then fine-tunes the network with RL. Using a single rollout of the policy, Monte-Carlo value estimation calculates the ground-truth values for both supervised learning and RL. 

Deep V-Learning has demonstrated success in simulation and/or in the real world but still suffers from the following drawbacks: (1) Deep V-Learning assumes that the state transitions of all surrounding humans are known and well-defined, which are in fact highly stochastic and difficult to model; (2) Since the networks are pre-trained with supervised learning, they share the same disadvantages with the demonstration policy, which are hard to be corrected by RL; (3) Monte-Carlo value estimation is not scalable with increasing time horizon; and (4) To achieve the best performance, Deep V-Learning needs state information of all humans. If applied to real robots, a real-time human detector with a $360^{\circ}$ field of view is required, which can be expensive or impractical.

To tackle these problems, we introduce a policy network trained with model-free RL, which does not need state transitions, Monte-Carlo value estimation, or expert supervision. 
Further, we show that incorporating both spatial and temporal reasoning in our network improves performance in challenging navigation environments over prior methods.

\subsection{Spatio-temporal graphs and structural-RNN}
St-graph is a type of conditional random field~\cite{lafferty2001conditional} with wide applications~\cite{jain2016structural,yu2018spatio,fan2019understanding,khodayar2018spatio}. St-graphs use nodes to represent the problem components and edges to capture the spatio-temporal interactions~\cite{jain2016structural}. With each node or edge governed by a factor function, st-graph decomposes a complex problem into many smaller and simpler factors.

Jain \textit{et al} propose a general method called structural-RNN (S-RNN) that transforms any st-graph to a mixture of RNNs that learn the parameters of factor functions end-to-end~\cite{jain2016structural}. S-RNNs have been applied to research areas such as human tracking and human trajectory prediction~\cite{sadeghian2017tracking,vemula2018social}. However, the scope of these works is restricted to learning from static datasets. Applying st-graph to crowd navigation poses extra challenges in data collection and decision-making under uncertainty.
Although some works in crowd navigation have used graph convolutional network~\cite{kipf2016semi} to model the robot-crowd interactions~\cite{chen2019relational, chen2020robot_gaze}, to the best of our knowledge, our work is the first to combine S-RNN with model-free RL for robot crowd navigation.

\section{Methodology}
In this section, we first formulate the robot decision making in crowd navigation as an RL problem. Then, we present our approach to model crowd navigation scenario as an st-graph, which leads to the derivation of our DS-RNN network architecture.
\label{sec:methods}

\begin{figure*}[ht]
\centering
\includegraphics[scale=0.7]{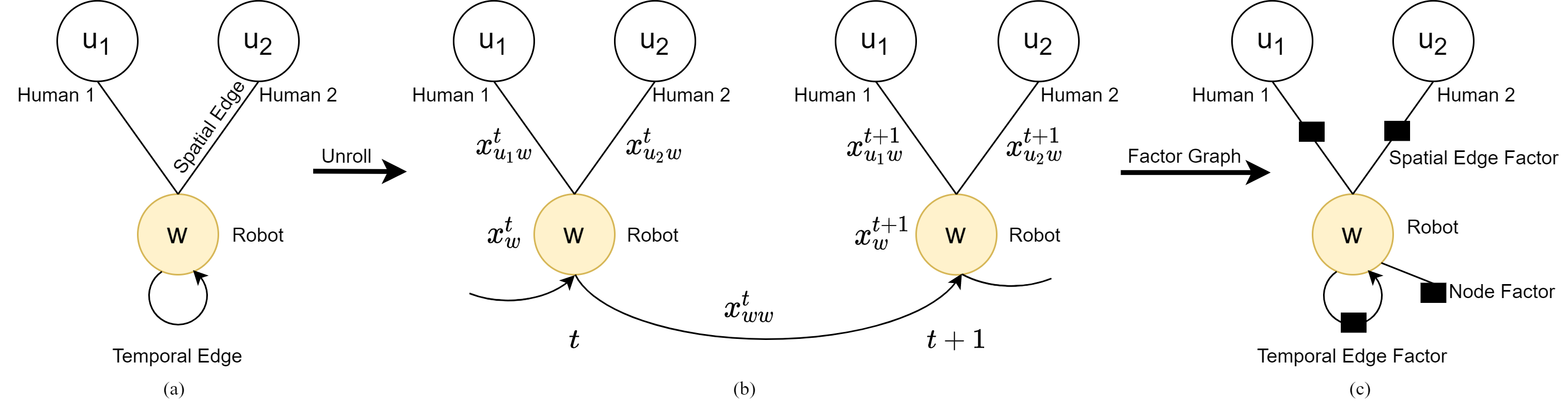}
\caption{\textbf{Conversion from the st-graph to the factor graph.} (a) St-graph representation of the crowd navigation scenario. We use $\mathrm{w}$ to denote the robot node and $\mathrm{u}_i$ to denote the $i$-th human node. (b) Unrolled st-graph for two timsteps. At timestep $t$, the node feature for the robot is $x^t_w$. The spatial edge feature between the $i$-th human and the robot is $x_{u_i w}^t$. The temporal edge feature for the robot is $x_{ww}^{t}$. (c) The corresponding factor graph. Factors are denoted by black boxes.}
\label{fig:st}
\vspace{-15pt}
\end{figure*}

\begin{figure}
\centering
\includegraphics[scale=0.5]{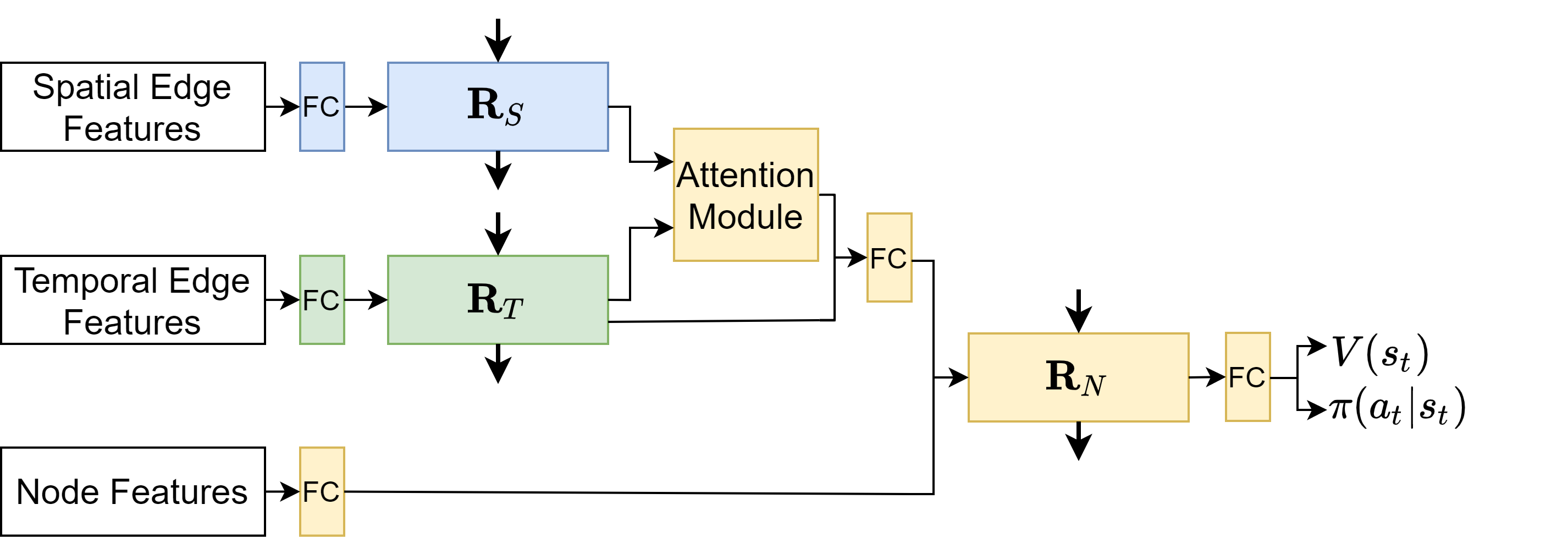}
\caption{\textbf{DS-RNN network architecture.} The components for processing spatial edge features, temporal edge features, and node features are in blue, green, and yellow, respectively. Fully connected layers are denoted as $FC$.}
\label{fig:network}
\vspace{-15pt}
\end{figure}

\subsection{Problem formulation}
Consider a robot interacting with an episodic environment with other humans. We model this interaction as an MDP, defined by the tuple $ \langle \mathcal{S}, \mathcal{A}, P, R, \gamma, \mathcal{S}_0 \rangle$. 
Suppose that all agents move in a 2D Euclidean space. Let $\mathbf{w}^t$ be the robot states and $\mathbf{u}_i^t$ be the $i$-th human's states observable by the robot. Then, the state $s_t \in \mathcal{S}$ for the MDP is $s_t=[\mathbf{w}^t, \mathbf{u}^t_1, ..., \mathbf{u}^t_n]$ assuming a total number of $n$ humans was involved at the timestep $t$. The robot state $\mathbf{w}^t$ consists of the robot's position $(p_x, p_y)$, velocity $(v_x, v_y)$, goal position $(g_x, g_y)$, maximum speed $v_{max}$, heading angle $\theta$, and radius $\rho$.
Each human state $\mathbf{u}_i^t$ consists of the human's position $(p_x^i, p_y^i)$. In contrast to previous works~\cite{van2011reciprocal,chen2017decentralized,chen2019crowd,chen2019relational},  the human state $\mathbf{u}_i^t$ does not include human's velocity and radius because they are hard to be measured accurately in the real world.

In each episode, the robot begins at an initial state $s_0\in \mathcal{S}_0$. At each timestep $t$, the robot takes an action $a_t\in\mathcal{A}$ according to its policy $\pi(a_t|s_t)$. In return, the robot receives a reward $r_t$ and transits to the next state $s_{t+1}$ according to an unknown state transition $P(\cdot|s_t, a_t)$. Meanwhile, all other humans also take actions according to their policies and move to the next states with unknown state transition probabilities. 
The process continues until $t$ exceeds the maximum episode length $T$, the robot reaches its goal, or the robot collides with any humans. 

Let $\gamma\in(0,1]$ be the discount factor. Then,  $R_t=\sum^\infty_{k=0}\gamma^{k}r_{t+k}$ is the total accumulated return from timestep $t$. The goal of the robot is to maximize the expected return from each state. The value of state $s$ under policy $\pi$, defined as $V(s)=\mathbb{E}[R_t|s_t=s]$, is the expected return for following policy $\pi$ from state $s$.

\subsection{Spatio-Temporal Graph Representation}
\label{sec:st_graph}
We formulate the crowd navigation scenario as a decentralized st-graph. Our graph $\mathcal{G} = (\mathcal{V}, \mathcal{E}_S, \mathcal{E}_T)$ consists of a set of nodes $\mathcal{V}$, a set of spatial edges $\mathcal{E}_S$, and a set of temporal edges $\mathcal{E}_T$. As shown in Fig.~\ref{fig:st}a, the nodes in the st-graph represent the agents, the spatial edges connect two different agents at the same timestep, and the temporal edges connect the same nodes at adjacent timesteps. We prune the edges and nodes not shown in Fig.~\ref{fig:st}a as they have little effect on the robot's decisions\footnote{From experiments, we find that a network derived from a full st-graph as in \cite{jain2016structural} performs very similarly to DS-RNN.}. The corresponding unrolled st-graph is shown in Fig.~\ref{fig:st}b. 

The factor graph representation of the st-graph factorizes the robot policy function into the robot node factor, spatial edge factors, and robot temporal edge factor. At each timestep, the factors take the node or edge features as inputs and collectively determine the robot's action. In Fig.~\ref{fig:st}c, factors are denoted by black boxes and have parameters that need to be learned.

We choose $x_{u_i w}^t$ to be the vector pointing from humans' position to the robot position, $(p^i_x - p_x, p^i_y - p_y)$, $x_{ww}^t$ to be robot velocity $(v_x, v_y)$, and $x^t_w$ to be $\mathbf{w}^t$. To reduce the number of parameters, all spatial edges share the same factor. This parameter sharing is important for the scalability of our st-graph because the number of parameters is kept constant with an increasing number of humans~\cite{jain2016structural}.

\subsection{Network Architecture}
As shown in Fig.~\ref{fig:network}, we derive our network architecture from the factor graph representation of the st-graph motivated by \cite{jain2016structural}. In our network, we represent each factor with an RNN, referred to as spatial edgeRNN $\mathbf{R}_S$, temporal edgeRNN $\mathbf{R}_T$, and nodeRNN $\mathbf{R}_N$ respectively. We use $W$ and $f$ to denote trainable weights and fully connected layers throughout this section.

The spatial edgeRNN $\mathbf{R}_S$ captures the spatial interactions between humans and the robot. $\mathbf{R}_S$ first applies a non-linear transformation to each spatial feature $x_{u_i w}^t$ and then feeds the transformed results to the RNN cell:
\begin{equation}
	h^t_{u_i w}=\mathrm{RNN}\left(h^{t-1}_{u_i w}, f_{\textrm{spatial}}(x_{u_i w}^t)\right )
\end{equation}
where $h^t_{u_i w}$ is the hidden state of the RNN at time $t$ for $i$-th human and the robot. Due to the parameter sharing mentioned in Section~\ref{sec:st_graph}, the spatial edge features between all human-robot pairs are fed into the same spatial edgeRNN. 

The temporal edgeRNN $\mathbf{R}_T$ captures the dynamics of the robot’s own trajectory. Similar to $\mathbf{R}_S$, $\mathbf{R}_T$ applies a linear transformation to the temporal edge feature and processes the results with its RNN cell:
\begin{equation}
	h^t_{ww}=\mathrm{RNN}\left(h^{t-1}_{ww},  f_{\textrm{temporal}}(x_{w w}^t)\right)
\end{equation}
where $h^t_{ww}$ is the hidden state of the RNN at time $t$.

The outputs of two edgeRNNs are fed into an attention module which assigns attention weights to each spatial edge. The attention mechanism is similar to the \textit{scaled dot product attention} in \cite{vaswani2017attention}. Let $V^t$ be the output of $\mathbf{R}_S$ at time $t$, $V^t=[h^t_{u_1 w},...,h^t_{u_n w}]^\top$, where $n$ is the number of spatial edges or humans.
Both  $V^t$ and $h^t_{ww}$ are first put through linear transformations:  
\begin{equation}
	Q^t=V^tW_{Q}, \quad K^t=h^t_{ww}W_{K}
\end{equation}
where $Q^t \in \mathbb{R}^{n \times d_k}$, $K^t \in \mathbb{R}^{1 \times d_k}$, and $d_k$ is a hyperparameter for the attention size. The attention weight at time $t$, $\alpha^t$, is calculated as 
\begin{equation}
	\alpha^t=\textrm{softmax}\left(\frac{n}{\sqrt{d_k}}Q^t (K^t)^\top\right)
\end{equation}
The output of the attention module at time $t$, $v^t_{\textrm{att}}$, is the weighted sum of spatial edges:
\begin{equation}
	v^t_{\textrm{att}}=(V^t)^\top \alpha^t
\end{equation}
The nodeRNN $\mathbf{R}_N$ uses the robot state, $x^t_w$, the weighted hidden states of $\mathbf{R}_S$, $v^t_{\textrm{att}}$, and the hidden states of temporal edgeRNN, $h^t_{ww}$, to determine the robot action and state value at each time $t$. The nodeRNN concatenates $v^t_{\textrm{att}}$ and $h^t_{ww}$ and embeds the concatenated results and the robot state with linear transformations:
\begin{equation}
	e^t=f_{\textrm{edge}}([v^t_{\textrm{att}}, h^t_{ww}]), \quad 
    n^t=f_{\textrm{node}}(x^t_w)
\end{equation}
Both $e^t$ and $n^t$ are concatenated and fed into $\mathbf{R}_N$ to get the nodeRNN hidden state. 
\begin{equation}
	h^t_{w}=\mathrm{RNN}\left(h^{t-1}_{w}, [e^t, n^t]\right)
\end{equation}
Finally, the $h^t_{w}$ is input to a fully connected layer to obtain the value $V(s_t)$ and the policy $\pi(a_t|s_t)$. We use Proximal Policy Optimization (PPO), a model-free policy gradient algorithm, for policy and value function learning~\cite{schulman2017proximal} and we adopt the PPO implementation from~\cite{pytorchrl}. To accelerate and stabilize training, we run twelve instances of the environment in parallel for collecting the robot's experiences. At each policy update, 30 steps of six episodes are used. 

By identifying the independent components of robot crowd navigation, we split the complex problem into smaller factors, and use three RNNs to efficiently learn the parameters of the corresponding factors. By combining all components above, the end-to-end trainable DS-RNN network performs spatial and temporal reasoning to determine the robot action. 

\section{Simulation Experiments}
\label{sec:sim_exp}

In this section, we describe the simulation environment for training and present our experimental results in simulation. 
\subsection{Simulation environment}
Fig.~\ref{fig:env} shows our 2D simulation environments adapted from \cite{chen2019crowd}. We use holonomic kinematics for each agent, whose action at time $t$ consists of the desired velocity along the $x$ and $y$ axis, $a_t=[v_x, v_y]$. All humans are controlled by ORCA with randomized maximum speed and radius. 
We assume that 
humans react only to other humans but not to the robot. This invisible setting prevents our model from learning an extremely aggressive policy in which the robot forces all humans to yield while achieving a high reward. 
We also assume that all agents can achieve the desired velocities immediately, and they will keep moving with these velocities for $\Delta t$ seconds. 
We define the update rule for an agent's position $p_x$, $p_y$ as follows:
\begin{equation}
\label{eqn:dynamics}
\begin{split}
    p_x[t+1] &=p_x[t] + v_x[t] \Delta t
    \\
    p_y[t+1] &=p_y[t] + v_y[t] \Delta t  
\end{split}
\end{equation}
\subsubsection{Environment configurations}
\begin{figure}
    \centering
    
    \begin{subfigure}{0.25\textwidth}
        
        \includegraphics[width=\linewidth]{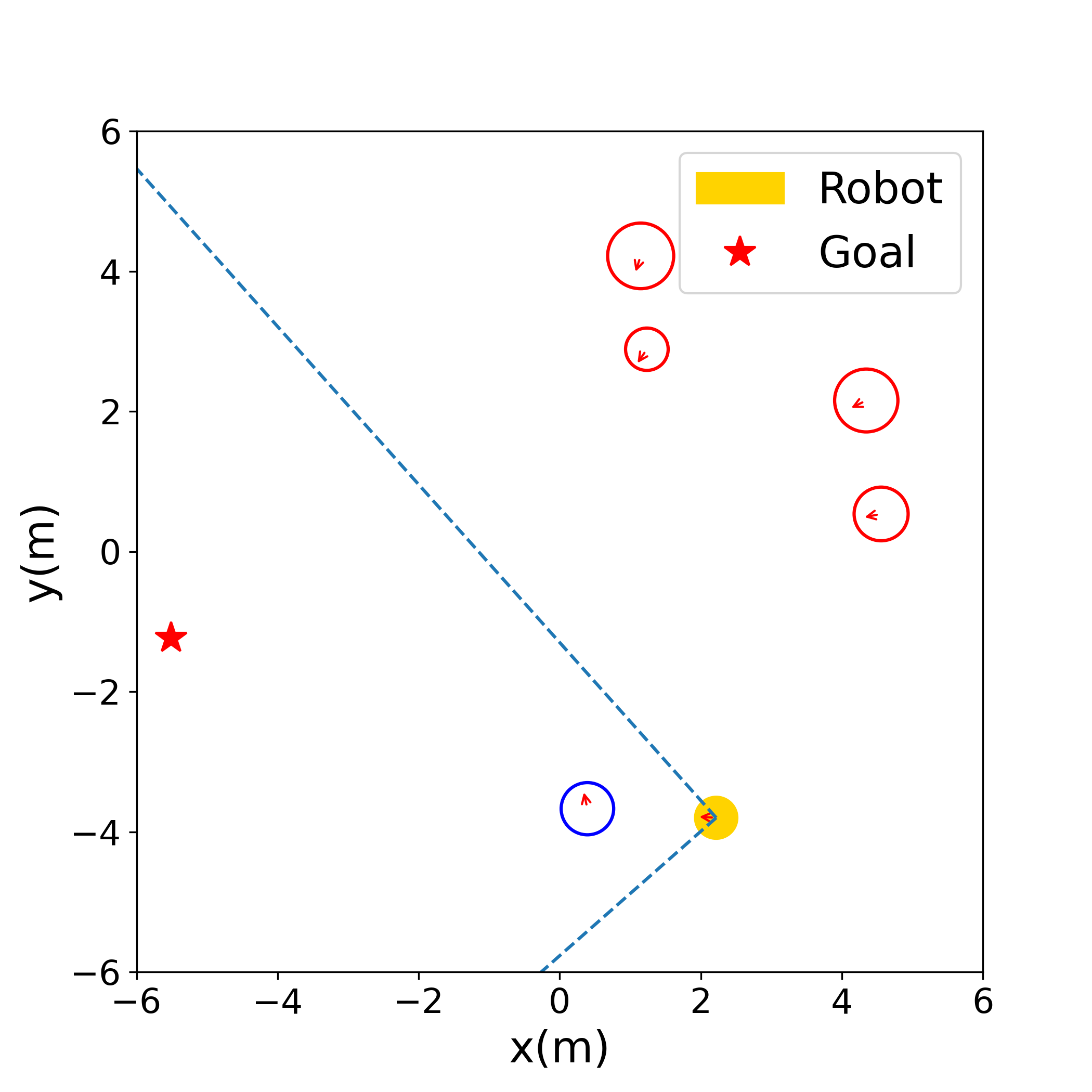}
        \caption{} \label{fig:enva}
    \end{subfigure}%
    \begin{subfigure}{0.25\textwidth}
        
        \includegraphics[width=\linewidth]{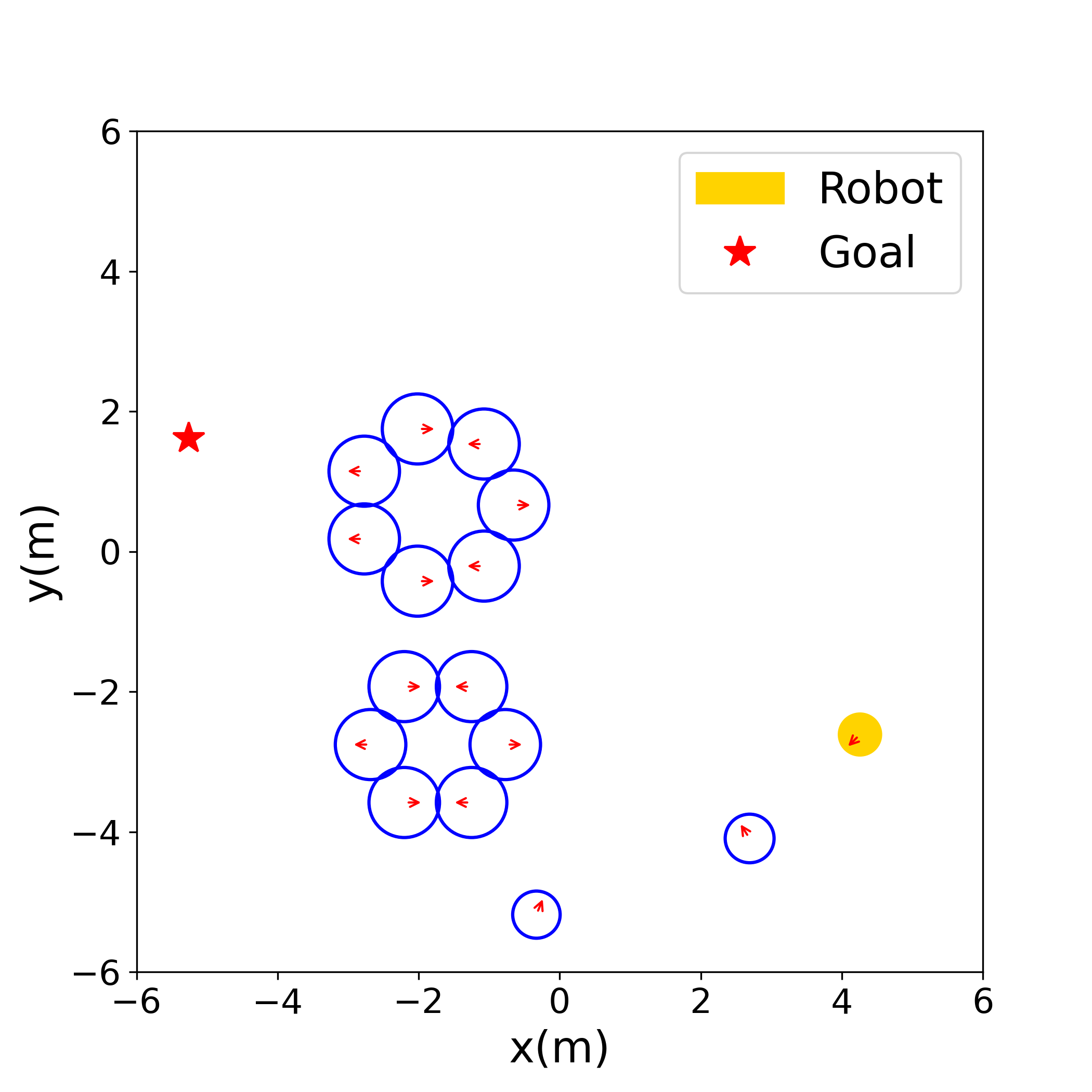}
        \caption{} \label{fig:envb}
    \end{subfigure}%

    \caption{\textbf{Illustration of our simulation environment.} 
    In a $12m\times12m$ $2D$ plane, the humans are represented as circles, the orientation of an agent is indicated by a red arrow, the robot is the yellow disk, and the robot's goal is the red star. We outline the borders of the robot FoV with dashed lines. The humans in the robot's FoV are blue and the humans outside are red.}
    \label{fig:env}
    \vspace{-15pt}
\end{figure}

Fig.~\ref{fig:enva} shows the FoV Environment, where the robot's field view (FoV) is within \ang{0} to \ang{360} and remains unchanged in each episode. The robot assumes that the humans out of its view proceed in a straight line with their last observed velocities. There are always five humans, whose starting and goal positions are randomly placed on a circle with radius $6m$. The FoV Environment simulates the limited sensor range of a robot, since deploying several sensors to obtain a \ang{360} FoV is usually expensive and unrealistic in the real world. 

Fig.~\ref{fig:envb} shows the Group Environment, where the robot's FoV is \ang{360} but the number of humans is large and remains the same in each episode. Among these humans, some form circle groups in random positions and do not move while the rest of them move freely. The Group Environment simulates the scene of a dense crowd with both static and dynamic obstacles. We use this environment to evaluate whether the robot policies have the \emph{freezing robot problem}.

\begin{figure*}[ht]
    \centering

    \includegraphics[scale=0.3]{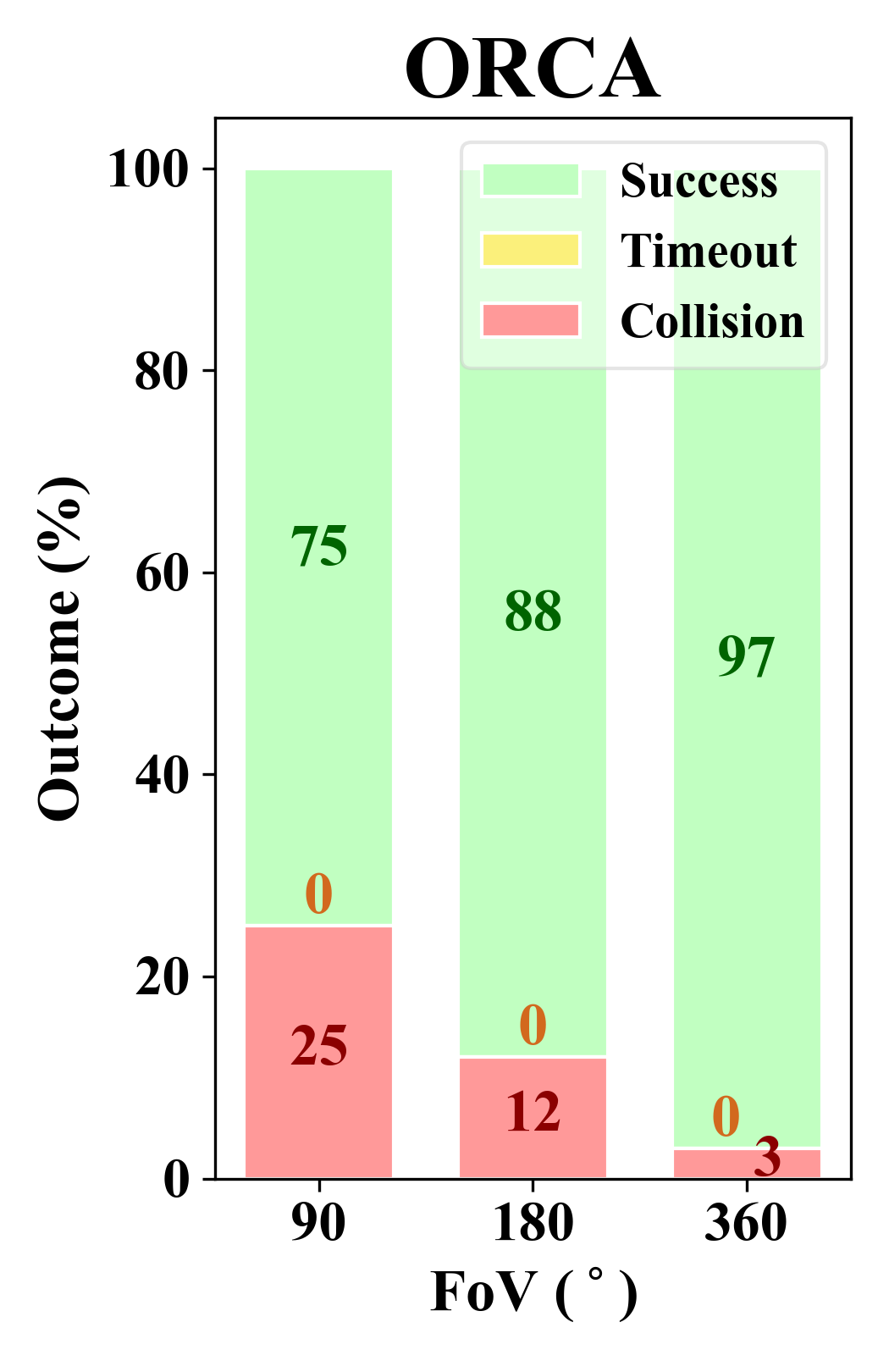}
    \includegraphics[scale=0.3,trim={1cm 0 0 0},clip]{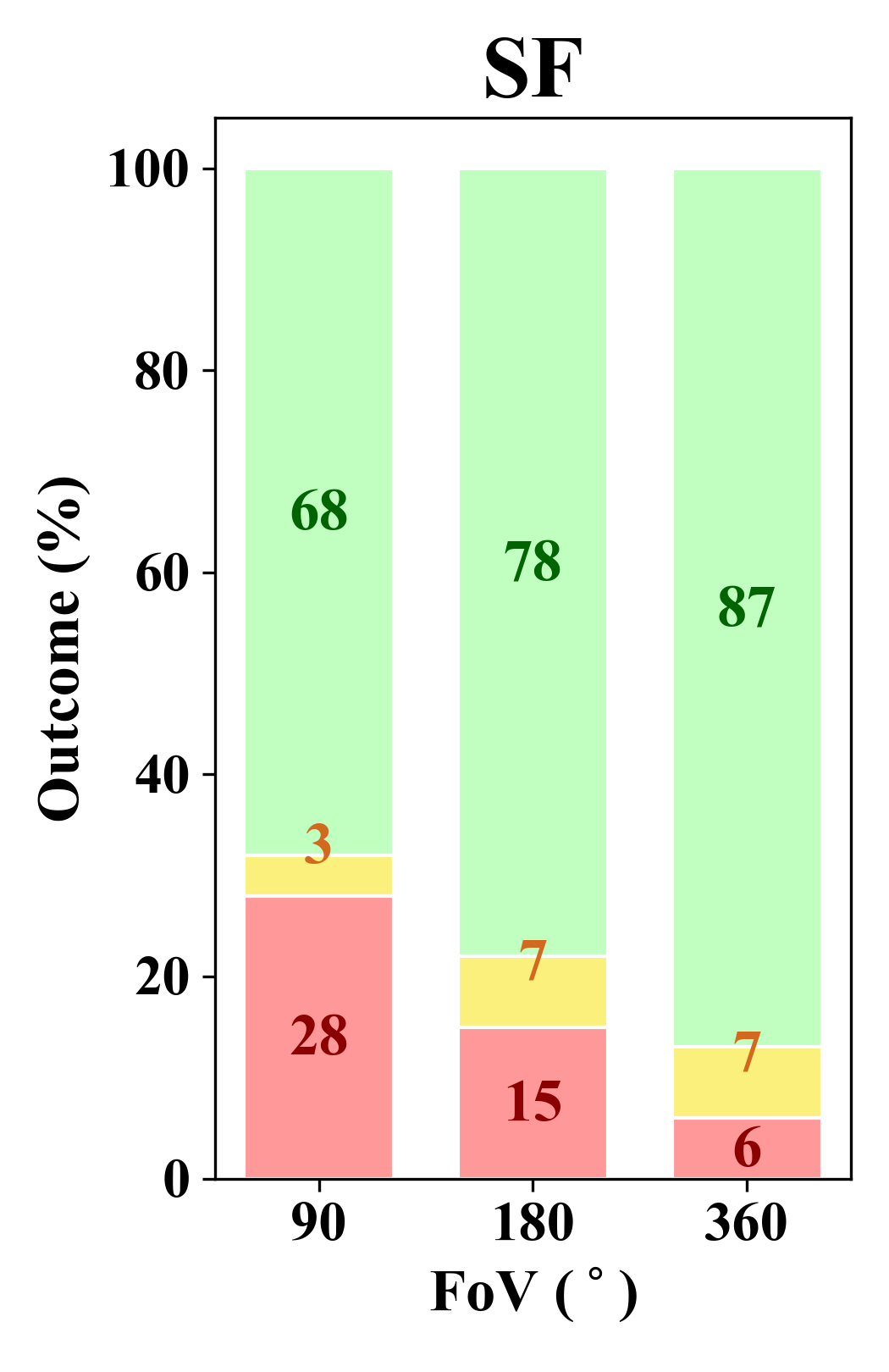}
    \includegraphics[scale=0.3,trim={1cm 0 0 0},clip]{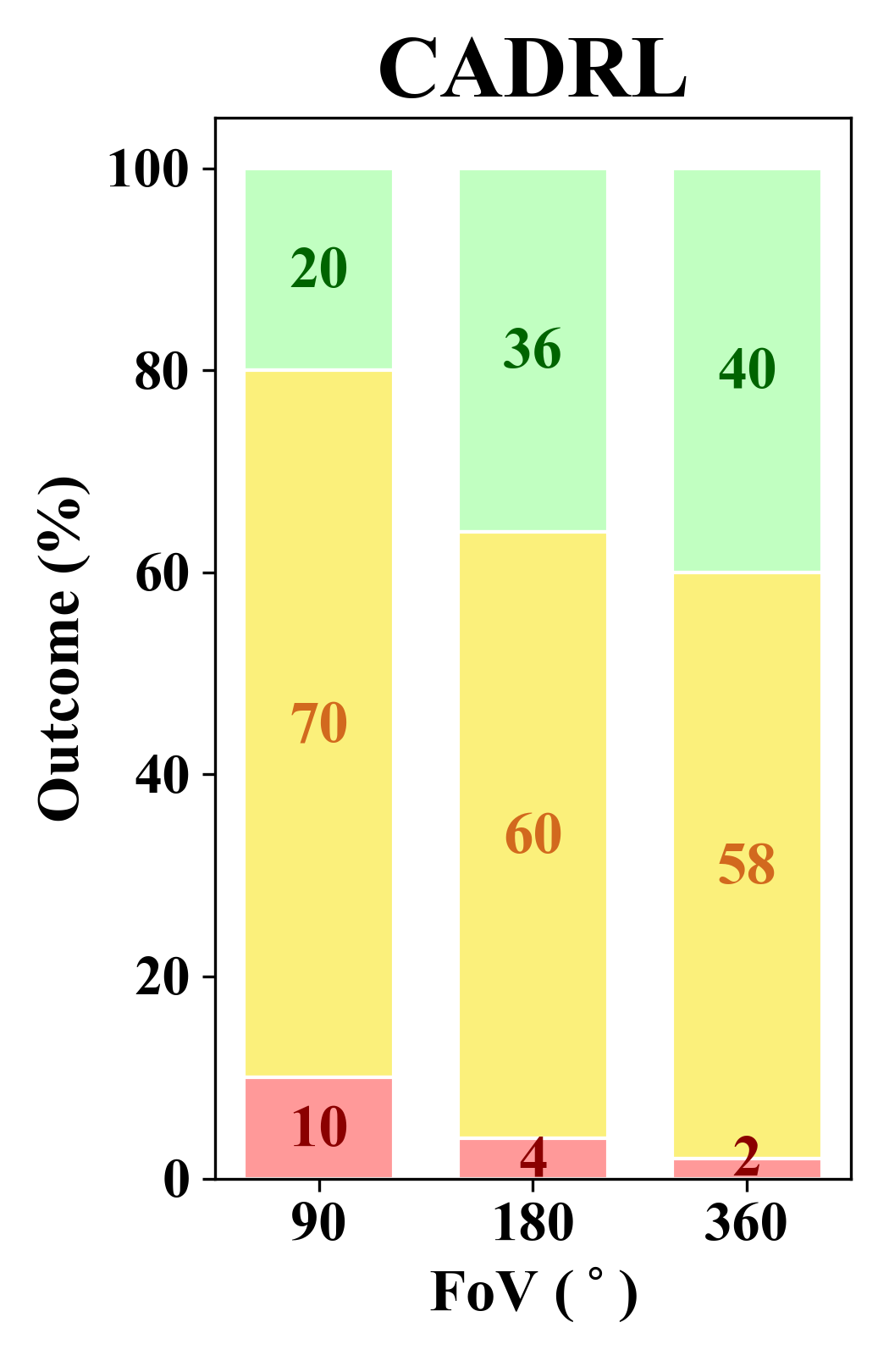}
    \includegraphics[scale=0.3,trim={1cm 0 0 0},clip]{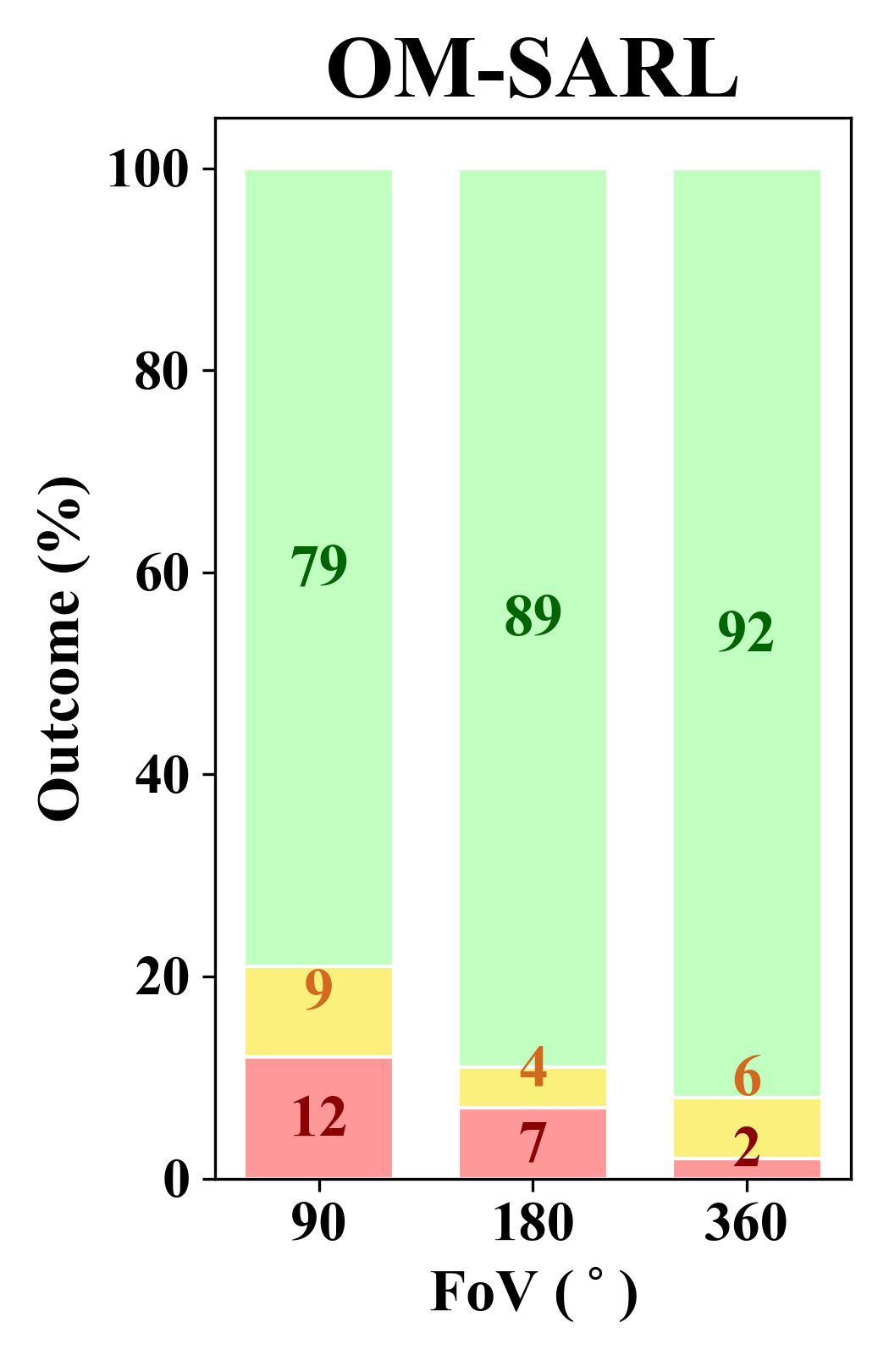}
    \includegraphics[scale=0.3,trim={1cm 0 0 0},clip]{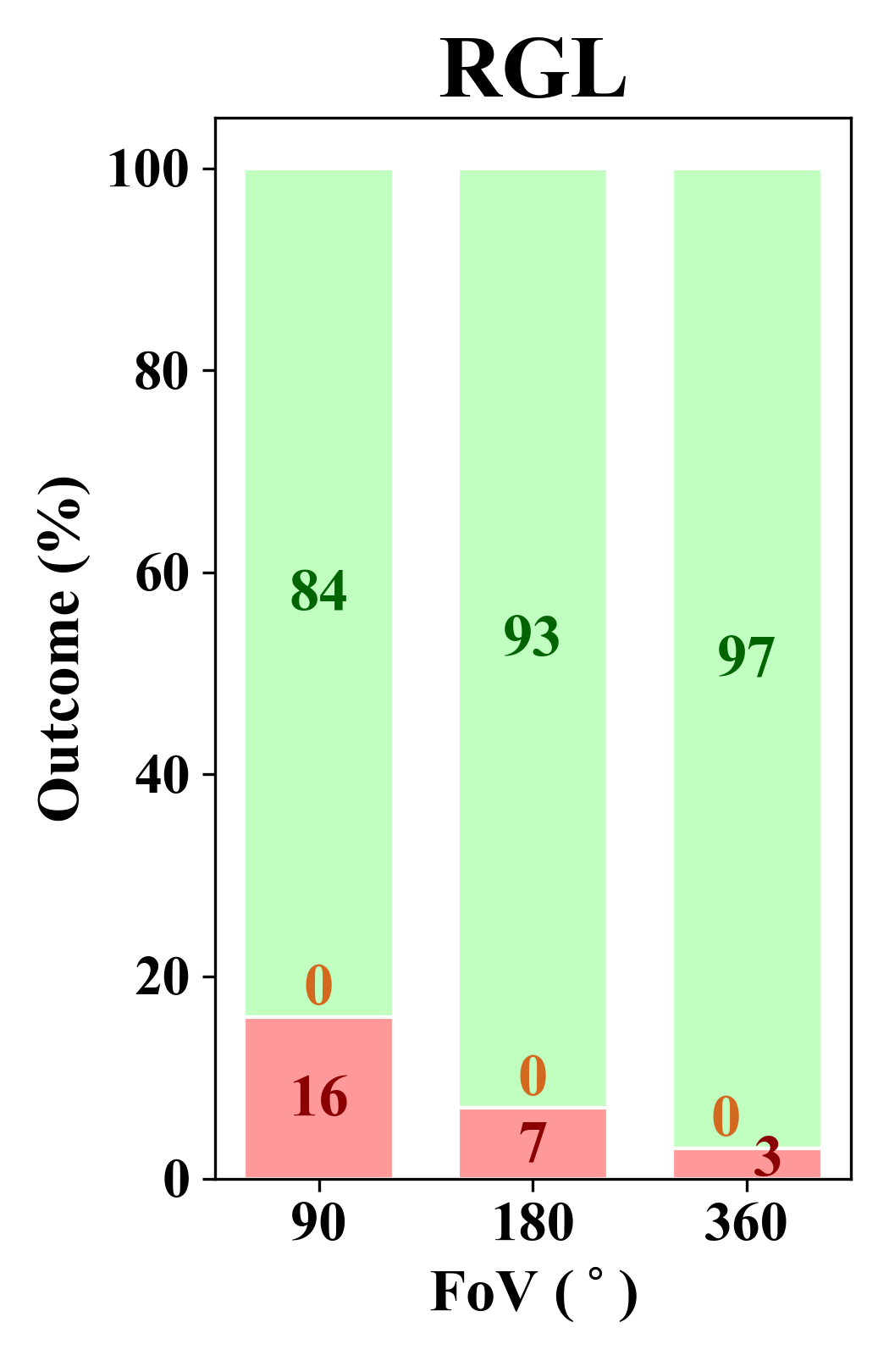}
    \includegraphics[scale=0.3,trim={1cm 0 0 0},clip]{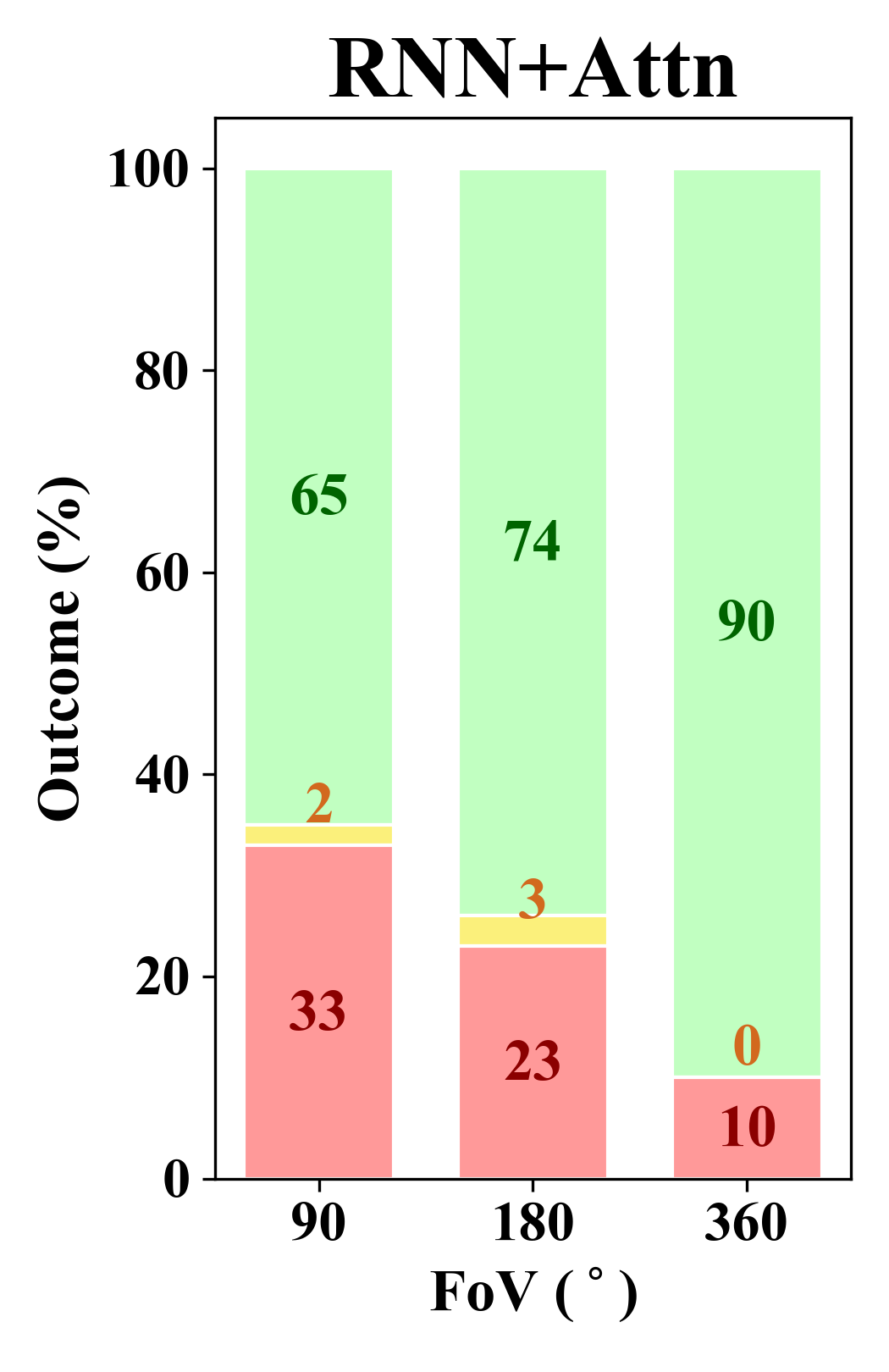}
    \includegraphics[scale=0.3,trim={1cm 0 0 0},clip]{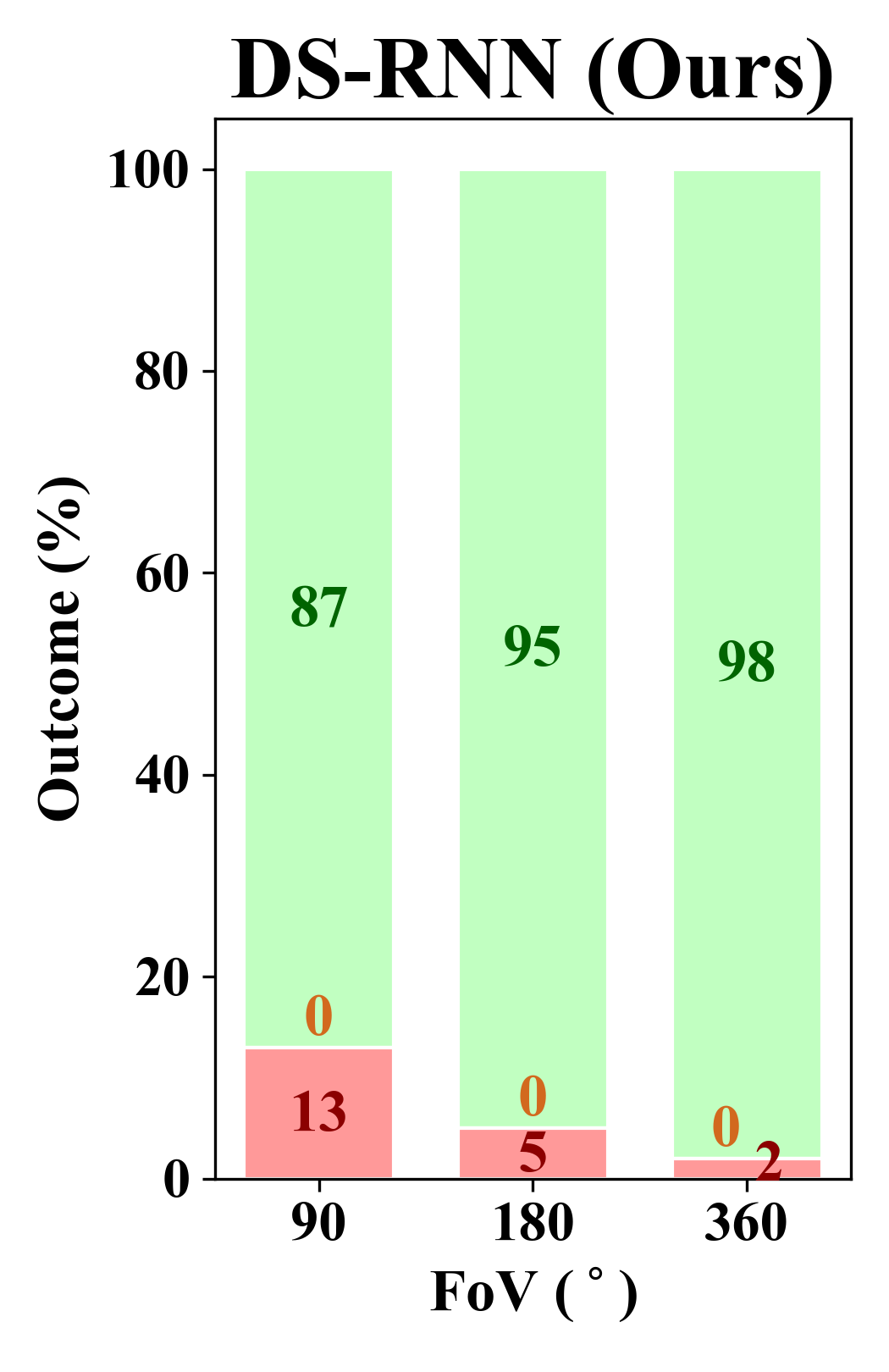}
    \caption{Success, timeout, and collision rates w.r.t. different FoV. The numbers on the bars indicate the percentages of the corresponding bars.}
    \label{fig:FOV_success}
    \vspace{-5pt}
\end{figure*}

\begin{figure*}[ht]
    \centering
    
    \includegraphics[scale=0.3]{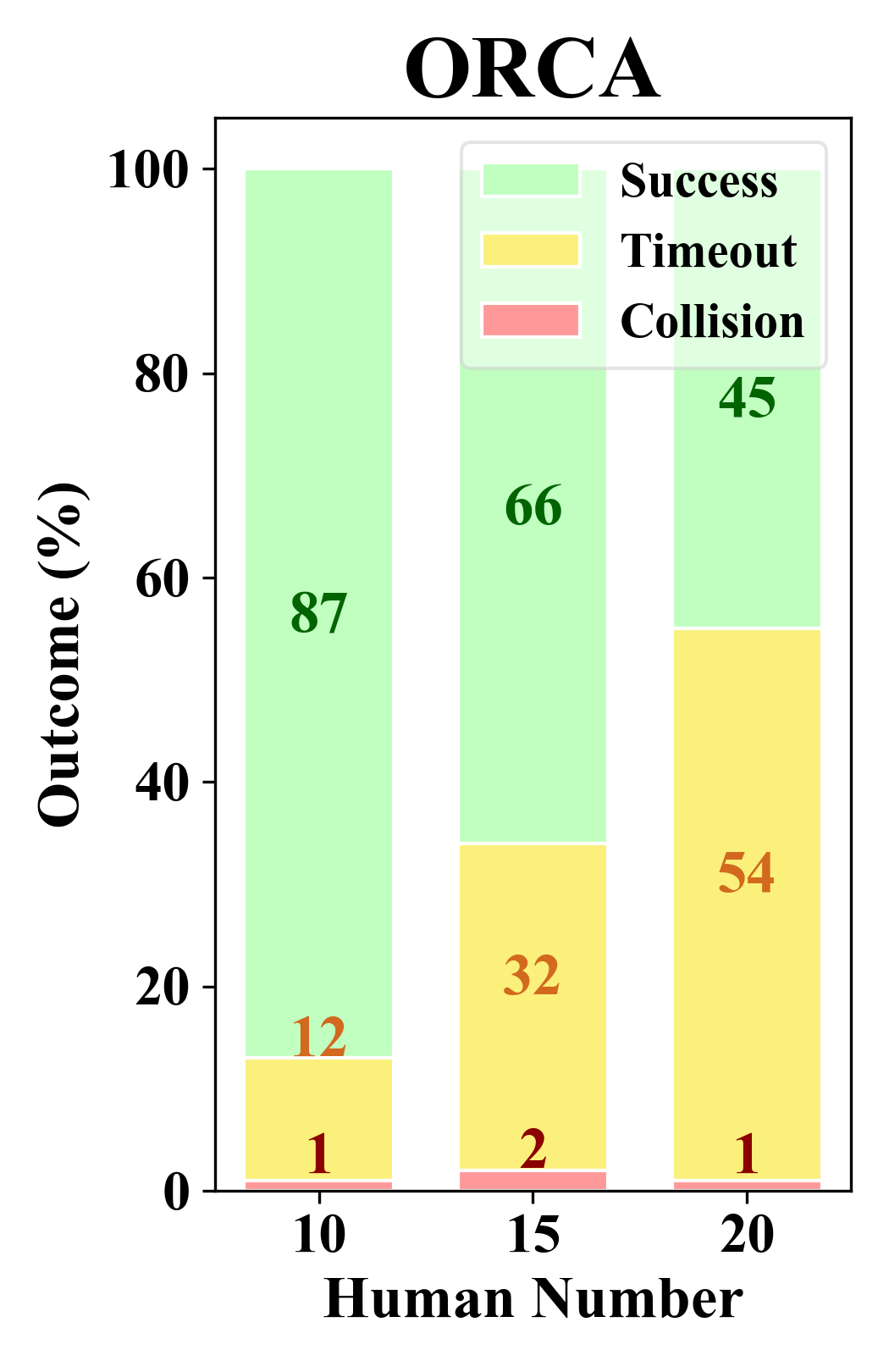}
    \includegraphics[scale=0.3,trim={1cm 0 0 0},clip]{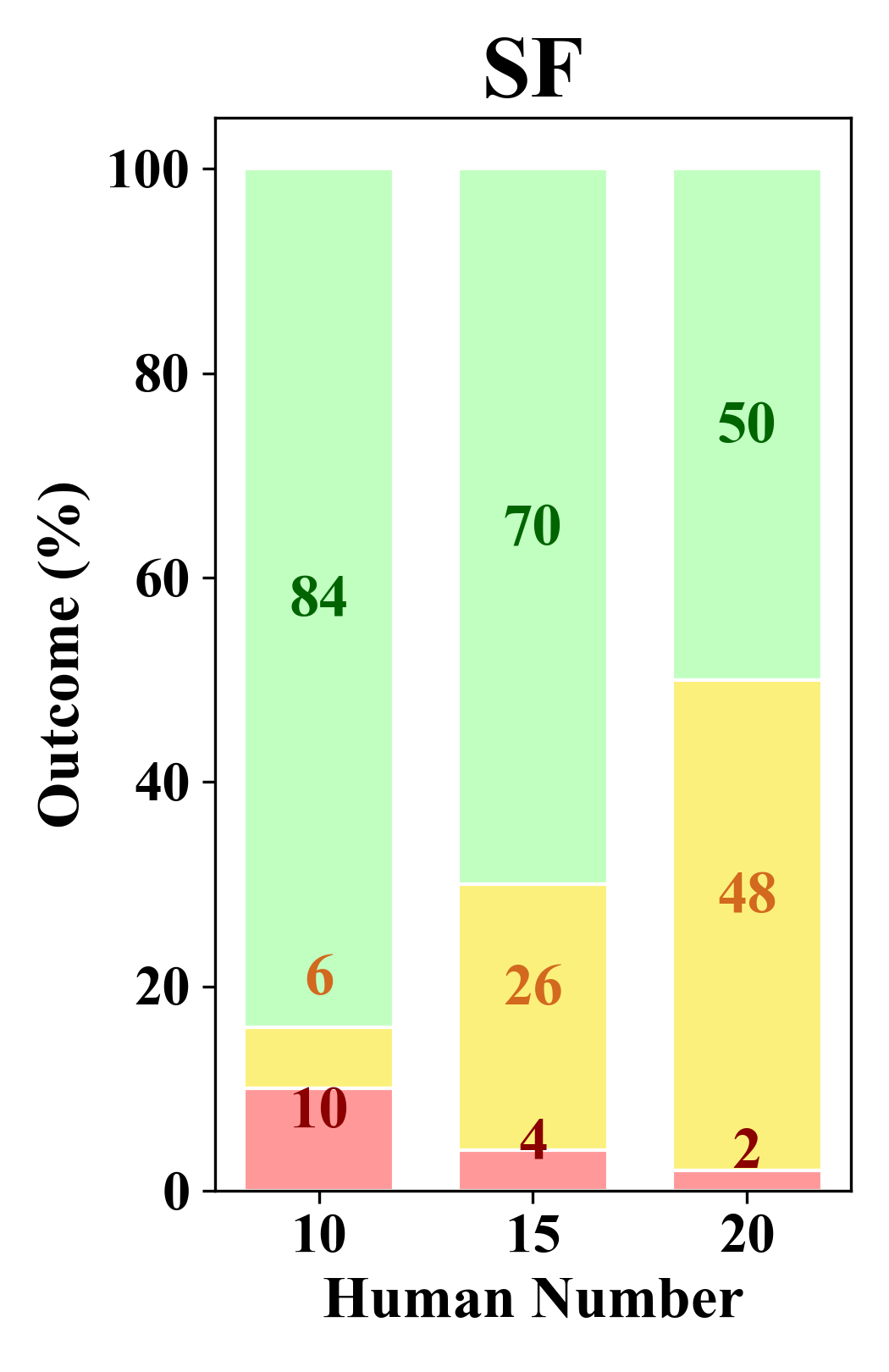}
    \includegraphics[scale=0.3,trim={1cm 0 0 0},clip]{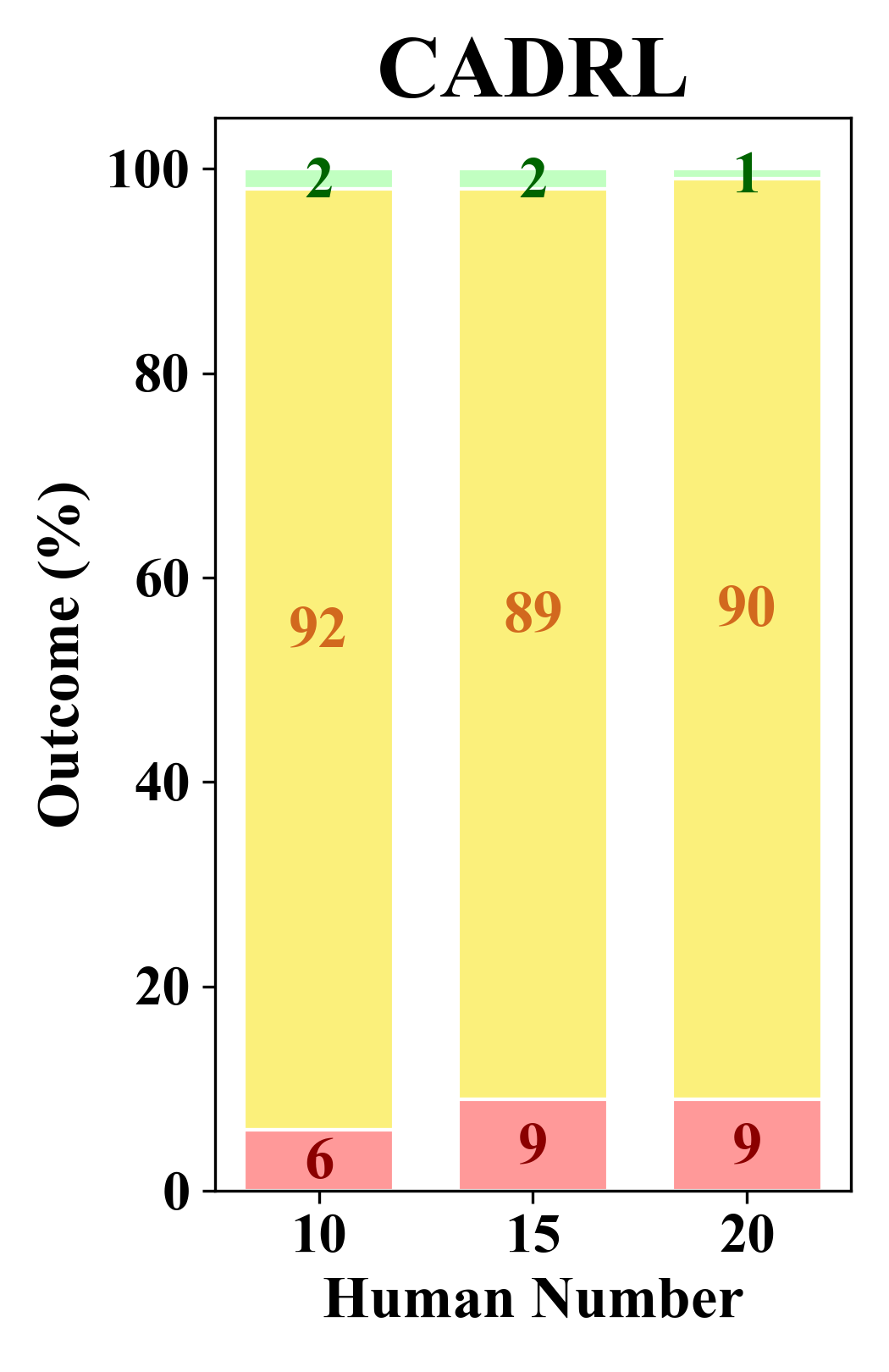}
    \includegraphics[scale=0.3,trim={1cm 0 0 0},clip]{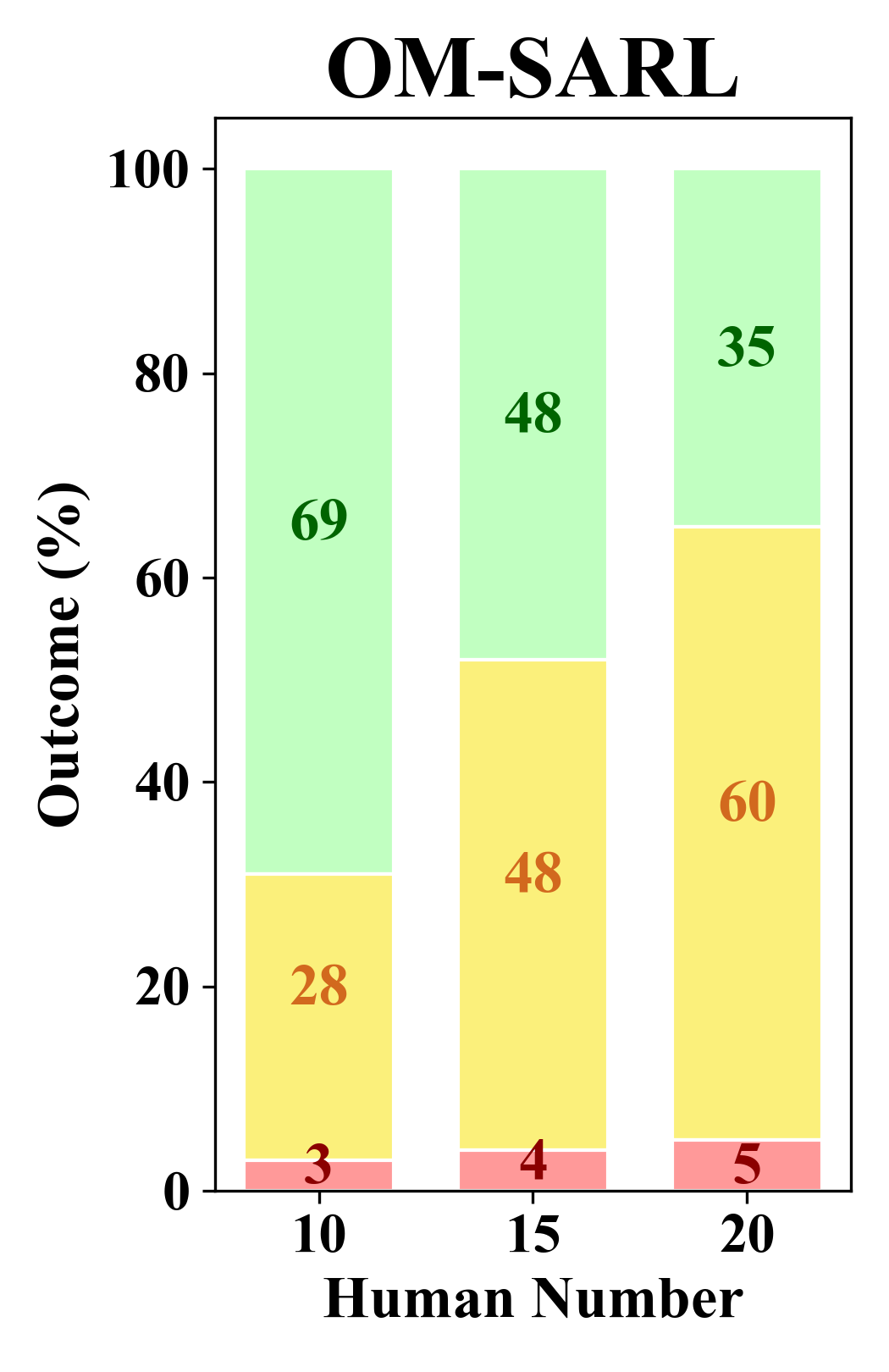}
    \includegraphics[scale=0.3,trim={1cm 0 0 0},clip]{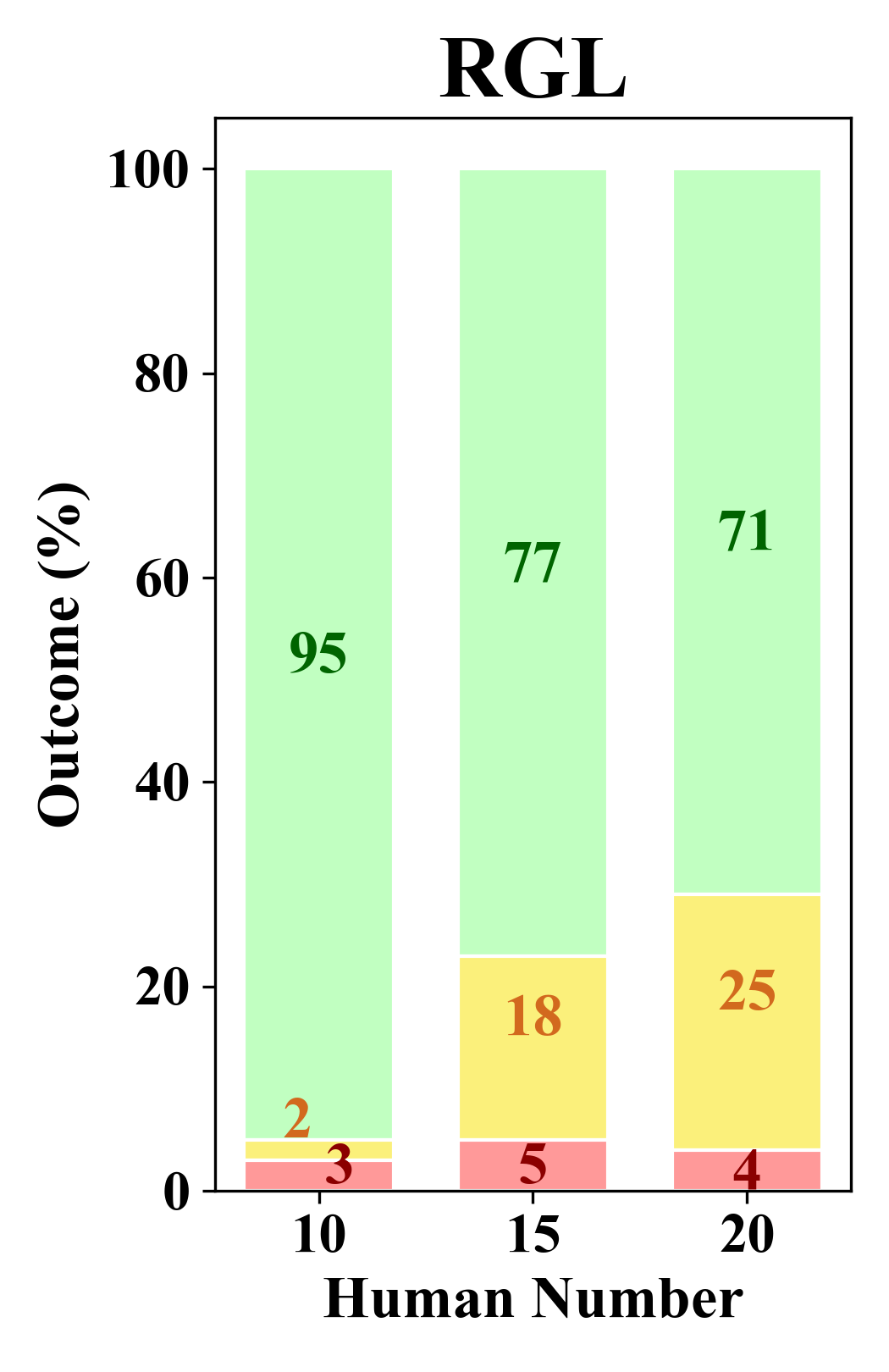}
    \includegraphics[scale=0.3,trim={1cm 0 0 0},clip]{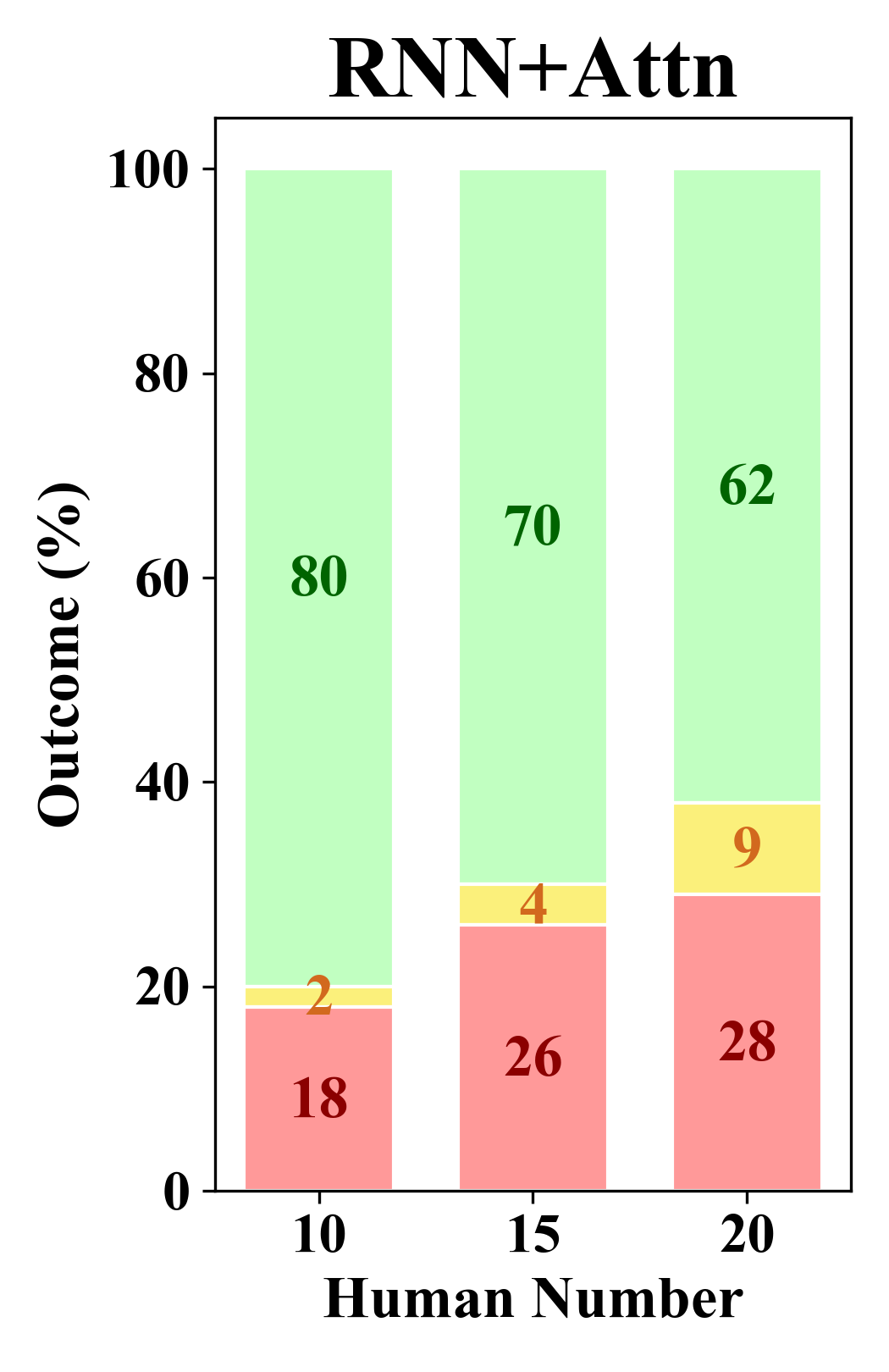}
    \includegraphics[scale=0.3,trim={1cm 0 0 0},clip]{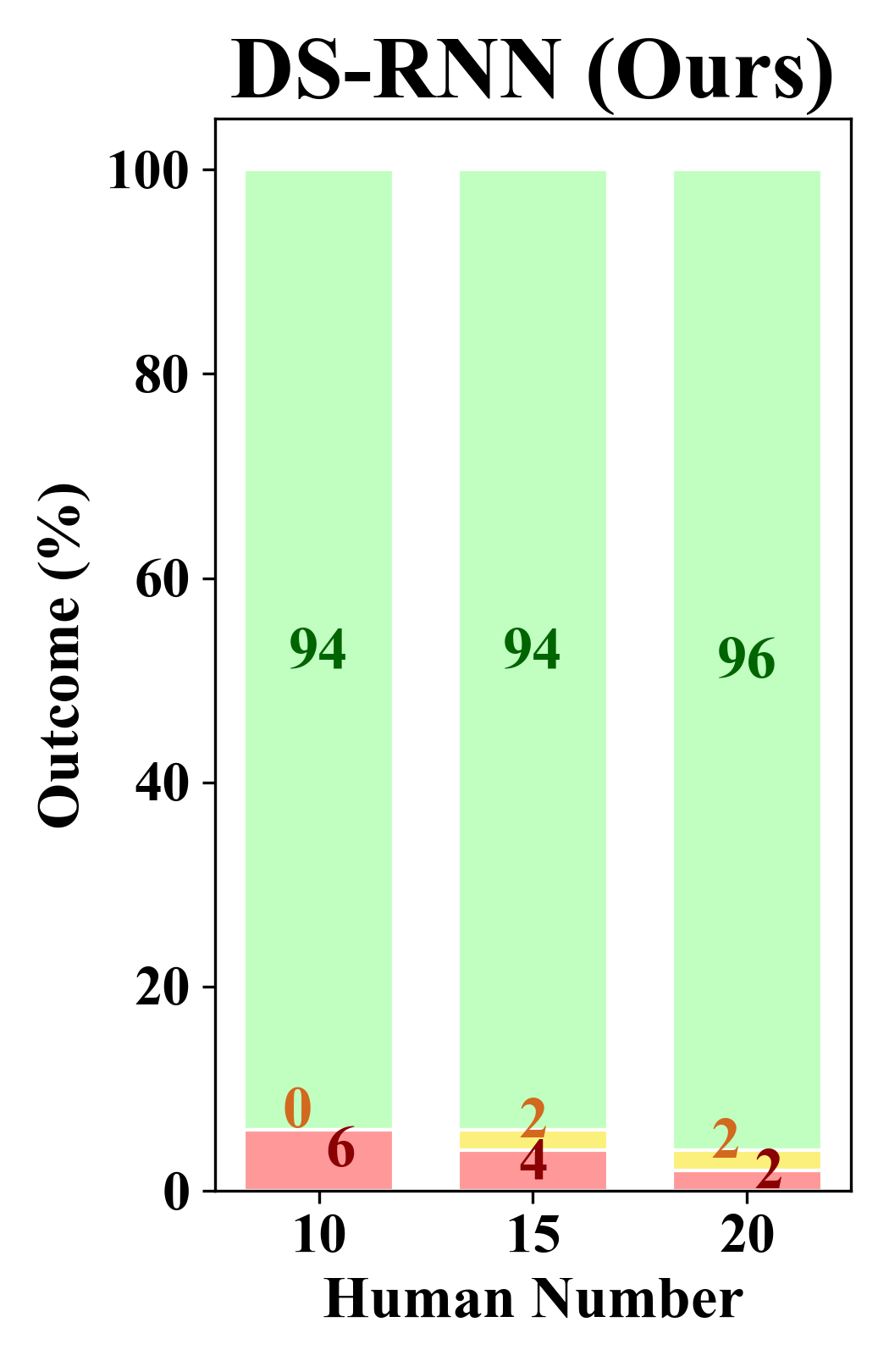}
    \caption{Success, timeout, and collision rates w.r.t. different number of humans.}
    \label{fig:group_success}
    \vspace{-10pt}
\end{figure*}

\begin{table}[ht]
  \begin{center}
    \caption{Navigation time (second) in two environments.}
    \label{tab:nav_time}
    \begin{tabular}{@{}llllclll@{}} 
    \toprule

     \multirow{2}{*}{\textbf{Method}} &  \multicolumn{3}{c}{\textbf{FoV}} & \phantom{abc} & \multicolumn{3}{c}{\textbf{Number of Humans}} \\
     \cmidrule{2-4} \cmidrule{6-8}
     & $\;$90$^{\circ}$  & $\;$180$^{\circ}$  & $\;$360$^{\circ}$  && $\;\;$10 & $\;\;$15 & $\;\;$20 \\ 
     \midrule
      ORCA & \textbf{9.12} & 9.96 & 10.40 && 15.94& 19.09&19.66 \\
      
      SF & 20.86 & 24.28 & 23.96 && 24.60 & 30.26 & 31.12 \\
      
      CADRL & 30.84 & 27.54 & 33.46 && 32.62& 36.81 & 41.75  \\
      
      OM-SARL & 18.32 & 13.70 & 21.04 && 27.25& 23.79& 29.24 \\
      
      RGL & 9.54 & \textbf{9.48} & \textbf{9.59} && \textbf{13.22} & \textbf{14.75} & 16.44 \\
     
      RNN+Attn & 16.57 & 14.00 & 10.96 && 16.01& 21.31& 25.55  \\
      
      DS-RNN  & 11.83 & 10.99 & 11.79 && 13.51& 15.64& \textbf{15.52}  \\
      
      \bottomrule
    \end{tabular}
  \end{center}
  \vspace{-20pt}
\end{table}

To simulate the variety of complexities in real-world crowd navigation, we add the following randomness and features not included in the original simulator from \cite{chen2019crowd}. When an episode begins, the robot's initial position and goal are chosen randomly. In addition, all humans occasionally change their goal positions within an episode. Finally, to simulate a continuous human flow, immediately after humans arrive at their goal positions, they will move to new random goals instead of remaining stationary at their initial destinations.

\subsubsection{Reward function}

The reward function awards the robot for reaching its goal and penalizes the robot for colliding with humans or getting too close to humans. In addition, we add a potential-based reward shaping to guide the robot to approach the goal:
\begin{equation}
\label{eq:reward}
\begin{split}
\begin{gathered}
    r(s_t, a_t)  = 
        \begin{cases}
            -20, & \text{if } d_{min}^t < 0\\
            2.5(d_{min}^t - 0.25), & \text{if } 0<d_{min}^t<0.25\\
            10, & \text{if } d_{goal}^t \leq \rho_{robot}\\
            2(-d_{goal}^t+d_{goal}^{t-1}), & \text{otherwise}.
        \end{cases}
\end{gathered}
\end{split}
\end{equation}
where $d_{min}^t$ is the minimum separation distance between the robot and any human at time $t$, and $d_{goal}^t$ is the $L2$ distance between the robot position and goal position at time $t$. 
Intuitively, the robot gets a high reward when it approaches the goal while maintaining a safe distance from all humans. 


\subsection{Experiment setup}


\subsubsection{Baselines and Ablation Models}
We compare the performance of our model with the representatives of the three types of methods in Section~\ref{sec:related}. We use ORCA and SF as the baselines for reaction-based methods; Relational Graph Learning (RGL)~\cite{chen2019relational} as a baseline for both trajectory-based methods and Deep V-Learning; and CADRL~\cite{chen2017decentralized} and OM-SARL~\cite{chen2019crowd} as the baselines for Deep V-Learning. 

To remove the performance gain caused by other factors such as model-free RL and RNN, we also implement an ablation model, called RNN+Attn, by adding an RNN to the end of OM-SARL network. For RNN+Attn, the attention module assigns attention weights on the state features of humans. The weighted human features are then concatenated with robot state features to form the joint state features which are passed to an RNN network with the same size and sequence length as the robot nodeRNN in our model. 
Both networks are trained using PPO with the same hyper-parameters and thus the results serve as a clean comparison to highlight the benefits of our DS-RNN. 

\subsubsection{Training} We use the same reward as defined in Equation~\ref{eq:reward} for CADRL, OM-SARL, RGL, RNN+Attn, and DS-RNN. 
The network architectures of all methods are kept the same in all experiments.
We train DS-RNN and RNN+Attn for $1\times 10^7$ timesteps with a learning rate $ 4\times 10^{-5}$. We train all baselines as stated in the original papers.

\subsubsection{Evaluation}
We evaluate the performance of all the models with six experiments: for the FoV Environment, the FoV of the robot is \ang{90}, \ang{180}, or \ang{360}; for the Group Environment, the number of humans is 10, 15, or 20. For each of the six experiments, we test all the models with $500$ random unseen test cases. We measure the percentage of success, collision, and timeout episodes, as well as the average navigation time of the successful episodes.



\subsection{Results}

\begin{figure}
    \centering
    Group Environment (20 Humans)
    \begin{subfigure}{0.16\textwidth}
        
        \includegraphics[width=\linewidth,trim={0.3cm 0 1.5cm 1.5cm},clip]{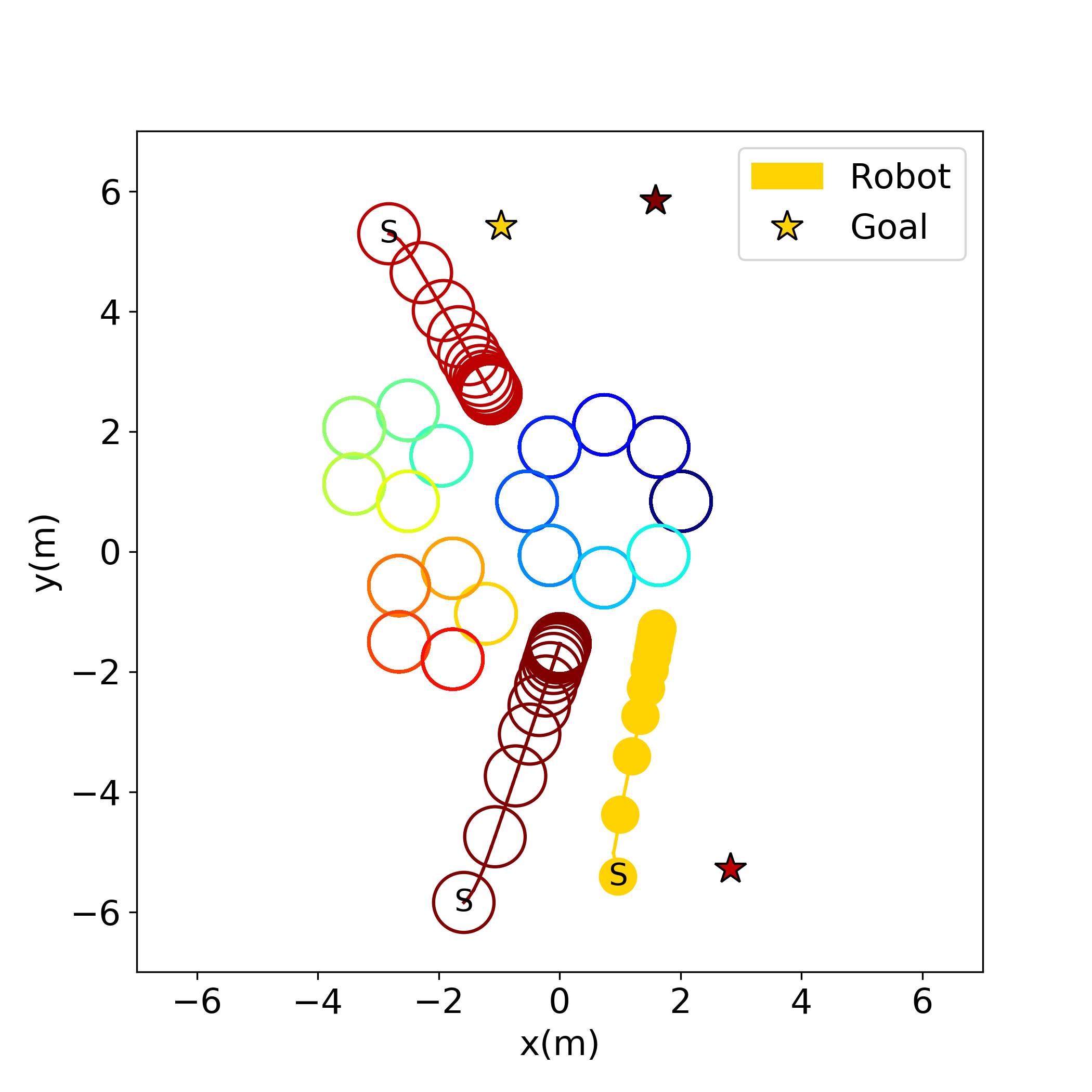}
        \caption{ORCA} 
    \end{subfigure}%
    \hspace*{\fill}
    \begin{subfigure}{0.16\textwidth}
        
        \includegraphics[width=\linewidth,trim={0.3cm 0 1.5cm 1.5cm},clip]{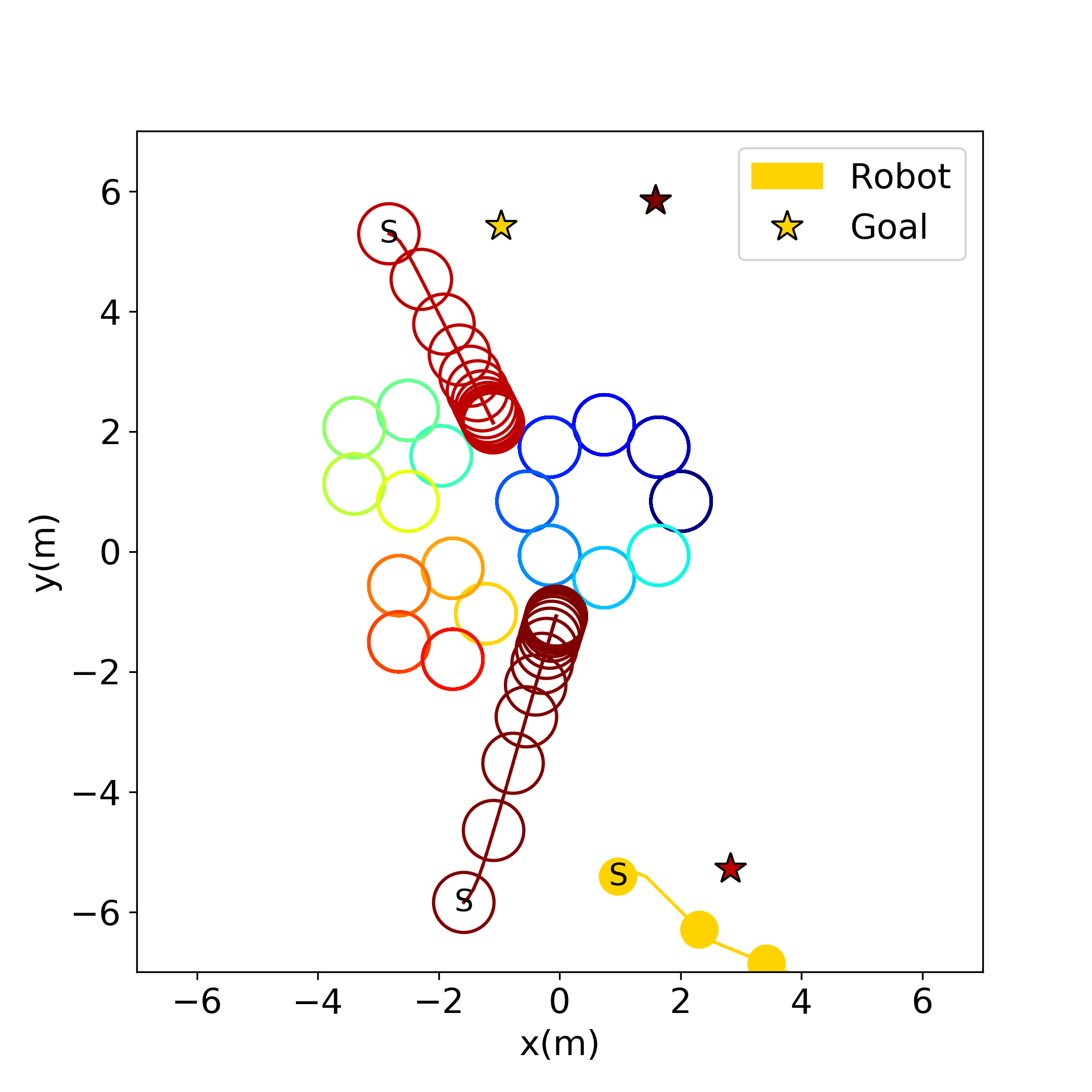}
        \caption{OM-SARL} 
    \end{subfigure}%
    \begin{subfigure}{0.16\textwidth}
        
        \includegraphics[width=\linewidth,trim={0.3cm 0 1.5cm 1.5cm},clip]{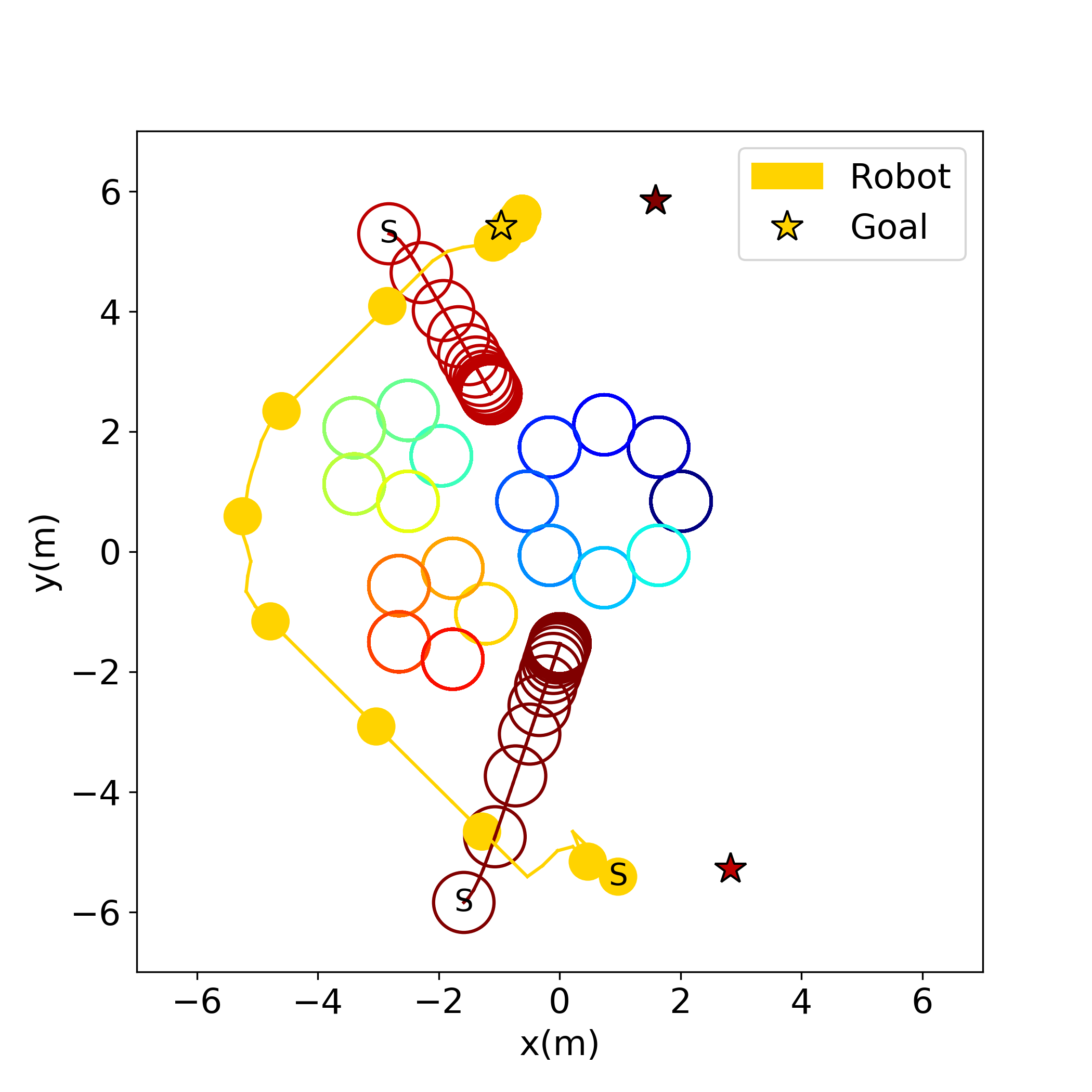}
        \caption{DS-RNN} 
    \end{subfigure}%

    \vspace{5pt}
    FoV Environment ($90^{\circ}$)
    \begin{subfigure}{0.16\textwidth}
        
        \includegraphics[width=\linewidth,trim={0.3cm 0 1.5cm 1.5cm},clip]{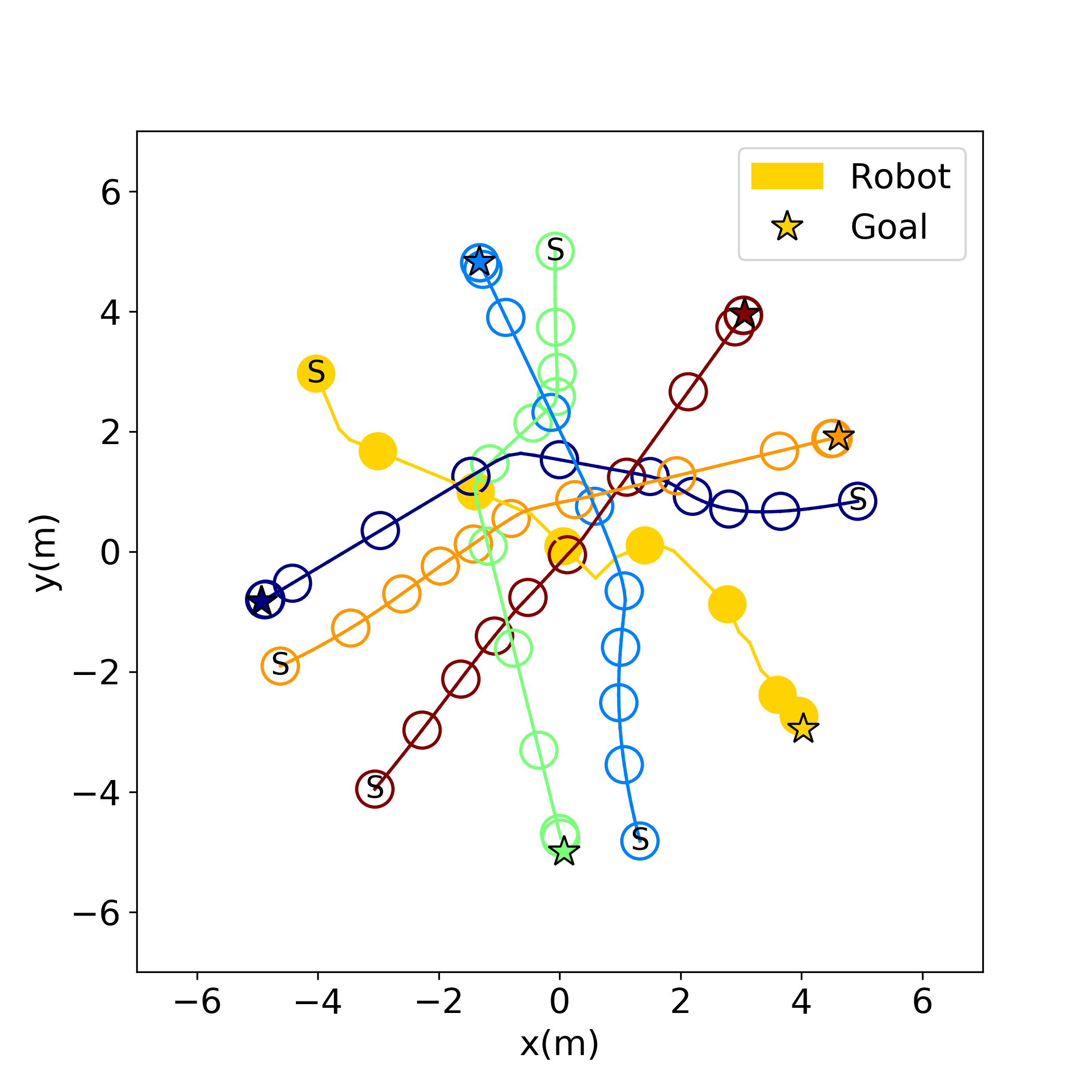}
        \caption{RGL} 
    \end{subfigure}%
    \begin{subfigure}{0.16\textwidth}
        
        \includegraphics[width=\linewidth,trim={0.3cm 0 1.5cm 1.5cm},clip]{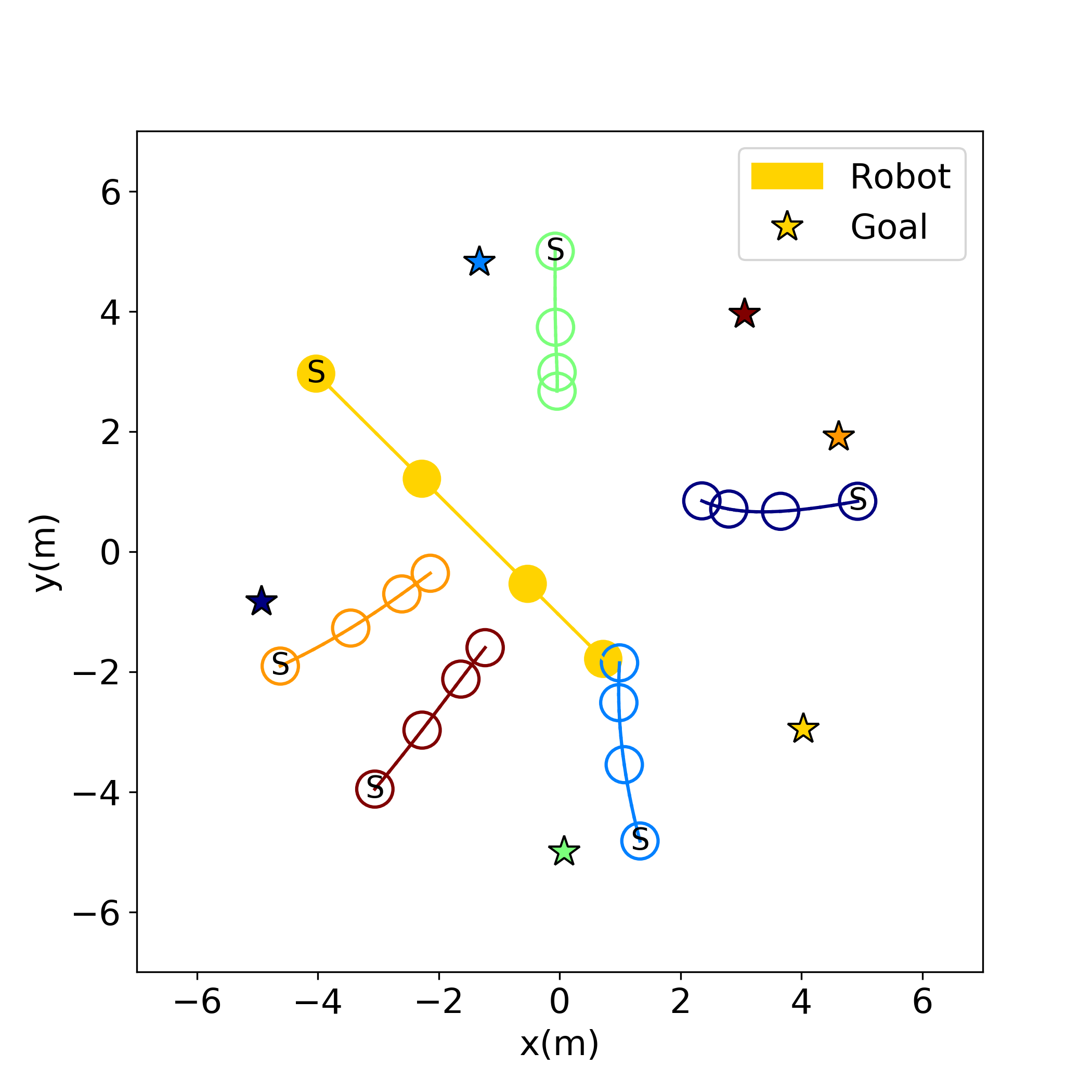}
        \caption{RNN+Attn} 
    \end{subfigure}%
     \begin{subfigure}{0.16\textwidth}
        
        \includegraphics[width=\linewidth,trim={0.3cm 0 1.5cm 1.5cm},clip]{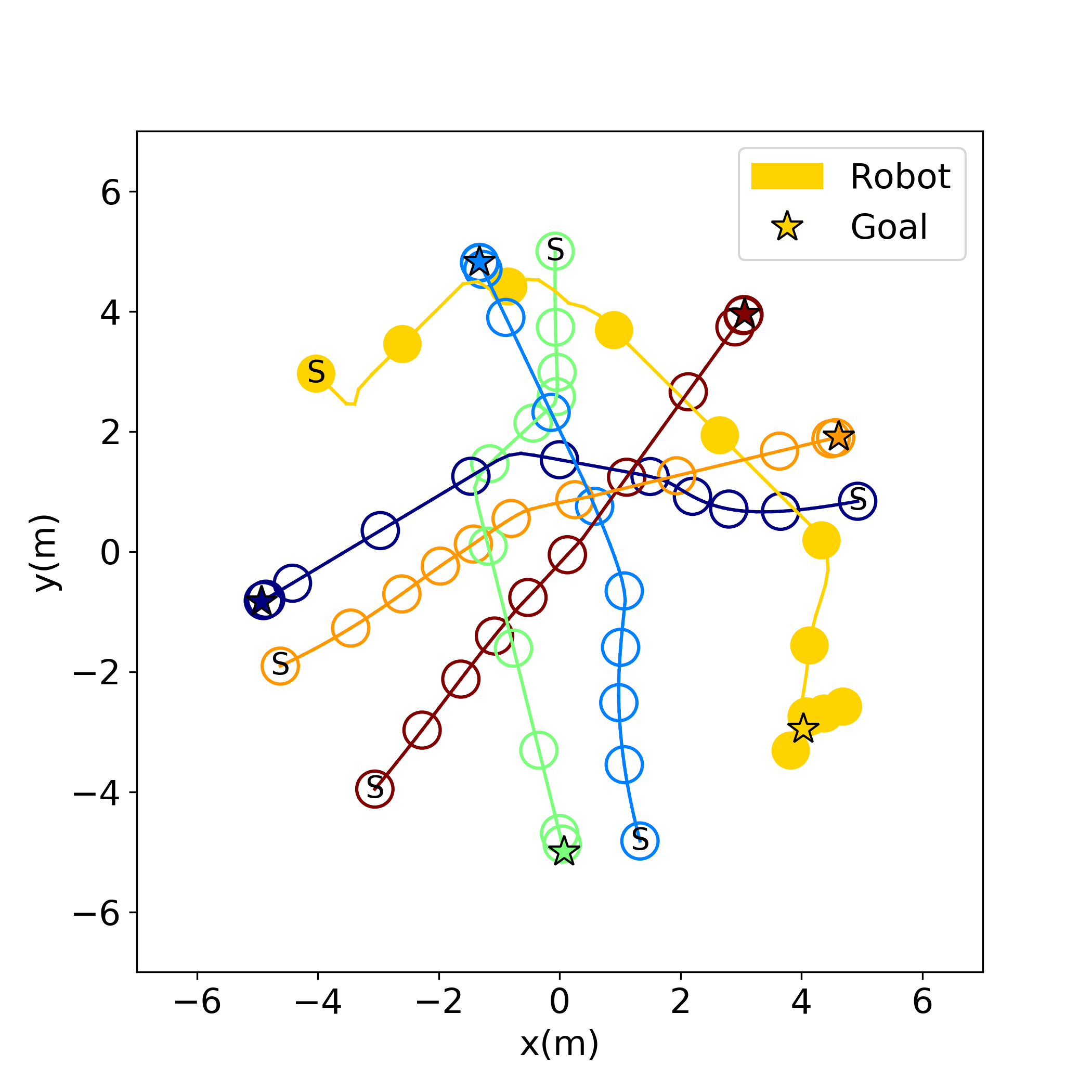}
        \caption{DS-RNN} 
    \end{subfigure}%
    \caption{\textbf{Trajectory comparisons of different methods with the same test cases.} Letter ``S'' denotes moving agents' starting positions, and stars denote moving agents' goals. The yellow filled circle denotes the robot. For the Group Environment (top), static humans are grouped in three circles.}
    \label{fig:traj}
    \vspace{-20pt}
\end{figure}
\subsubsection{Spatio-temporal reasoning}
We show the effectiveness of our DS-RNN architecture by comparing it with RNN+Attn. 
In Fig.~\ref{fig:FOV_success} and Fig.~\ref{fig:group_success}, compared with RNN+Attn, our model exhibits higher success rates and lower collision and timeout rates in all settings. In Table~\ref{tab:nav_time}, our model has shorter navigation time because DS-RNN often finds a better path (Fig.~\ref{fig:traj}e and \ref{fig:traj}f). 
We believe the main reason is that by formulating the crowd navigation into an st-graph, we decompose the robot decision making into smaller factors and feed each RNN with only relevant edge or node features. In this way, the three RNNs are able to learn their corresponding factors more effectively. By combining all factors (RNNs), the robot is able to explicitly reason about the spatial relationships with humans and its own dynamics to take actions. In contrast, RNN+Attn does not have such spatio-temporal reasoning and learns all factors with one single RNN, which explains its lower performance. 

\subsubsection{Comparison with traditional methods}
We compare the performance of our model with those of ORCA and SF. As shown in Fig.~\ref{fig:group_success}, in the Group Environment, ORCA and SF exhibit high timeout rates, which increases significantly as the number of humans increases. This observation indicates that the \emph{freezing robot problem} is prevalent in these mixed static and dynamic settings (Fig.~\ref{fig:traj}a). 
Also, as Table~\ref{tab:nav_time} suggests, the large navigation times show that both methods are overly conservative in dense crowds. 
In the FoV Environment, our method also outperforms ORCA and SF in most metrics, as shown in Fig.~\ref{fig:FOV_success} and Table~\ref{tab:nav_time}, because our method explores the environment and learns from the past experience during RL training. Combined with spatio-temporal reasoning, our method is able to better adapt to dense and partially observable environments. In addition, with our method, the robot is long-sighted because RL optimizes the policy over cumulative reward and the RNNs takes a sequence of trajectories to make decisions while ORCA and SF only consider the current state (Fig.~\ref{fig:traj}c).

\subsubsection{Comparison with Deep V-Learning}
We compare model-free RL training with Deep V-Learning used by CADRL, OM-SARL, and RGL. In Fig.~\ref{fig:group_success}, all three baselines exhibit large timeout rates. The reason is that the value networks are initialized by a suboptimal expert (ORCA) in supervised learning and are insufficient to provide good state value estimates, resulting in policies that inherit OCRA's drawbacks. 
In contrast, model-free RL enables RNN+Attn and DS-RNN to learn from scratch and prevents the network from converging to a suboptimal policy too early. 
In addition, despite the unknown state transitions of all agents, RNN+Attn and our method still perform better in all metrics compared with Deep V-Learning.
  
As shown in Fig.~\ref{fig:traj}d and \ref{fig:traj}f, RGL is competitive to our method in some cases, because the relational graph can perform spatial reasoning and human trajectory predictions make RGL long-sighted.
However, in RGL, the relational graph and the robot planner are separated modules, while our network is trained end-to-end and jointly learns the robot-human interactions and decision making.

\section{Real-world Experiments}
\label{sec:real_exp}
We evaluate our trained model’s performance on a TurtleBot 2i mobile platform as shown in Fig.~\ref{fig:opening}. 
An Intel RealSense depth camera D435 with an approximately \ang{69.4} FoV is used to obtain human positions. We use YOLOv3~\cite{yolov3} for human detection and Deep SORT~\cite{deepsort} for human tracking (our implementation is adopted from~\cite{deepsort_pytorch}). The human detection and tracking are combined with the camera depth information to calculate human positions.
An Intel RealSense tracking camera T265 is used to localize the robot and obtain the robot orientation. 
We run the above perception algorithms and our decision-making model on a remote host computer. The communication between the robot and the host computer is established by ROS. A video demonstration is available at \url{https://youtu.be/bYO-1IAjzgY}, where the robot successfully reaches the goals with maneuvers to maintain a safe distance with humans in various scenarios.
\section{Limitations}
\label{sec:limitation}
Our work encompasses the following limitations. First, the invisible setting of our simulation environment is different from the reality where the motions of pedestrians and those of the robot mutually affect each other. Since the robot does not affect human behaviors, it is difficult to quantify the social awareness of the robot and incorporate it into our design. Second, in the real world experiment, due to the sensor limitations, the detected human positions are noisy and thus cause differences in robot behaviors between the simulation and the real world. Third, the total number of humans is fixed for each network model, which poses challenges to the generalization of our model to navigation scenarios with real human flows.

\section{Conclusion and future work}
\label{sec:conclusion}
We propose a novel DS-RNN network that incorporates spatial and temporal reasoning into robot decision making for crowd navigation. 
We train our DS-RNN with model-free deep RL without any supervised learning or assumptions on agents' dynamics. 
Our experiments shows that our model outperforms various baselines in challenging simulation environments and show promising results in the real world.
Possible directions to explore in future work include (1) utilizing mutual interactions between the robot and humans to improve our model, and (2) enabling our network to take raw camera images as inputs to simplify detection and localization in the real world.

\section*{Acknowledgements}
We thank Zhengguan Dai for setting up the baseline code and Zhe Huang for feedback on paper drafts.



\bibliographystyle{IEEEtran}
\bibliography{BibFile}
\clearpage
\end{document}